\documentclass[journal]{IEEEtran}

\usepackage{graphicx}
\usepackage{amsmath,amssymb}
\usepackage{bm}
\usepackage[usenames, dvipsnames]{color}
\usepackage{tabularx}
\usepackage{soul}

\begin{document}

\title{Data-Driven Background Subtraction Algorithm for in-Camera Acceleration in Thermal Imagery}

\author{Konstantinos Makantasis, 
        Antonis Nikitakis,
        Anastasios Doulamis,
        Nikolaos Doulamis,
        and~ Yannis Papaefstathiou	
\thanks{K. Makantasis, A. Nikitakis and Y. Papaefstathiou with the Technical University of Crete, Chania, 73100 Greece e-mail: kmakantasis@isc.tuc.gr, a.s.nikitakis@gmail.com, ygp@mhl.tuc.gr}
\thanks{A. Doulamis and N. Doulamis are with the National Technical University of Athens, Athens, 15780 Greece e-mail: adoulam@cs.ntua.gr, ndoulam@cs.ntua.gr}
\thanks{The research leading to these results has received funding by the H2020 project TERPSICHORE funded by EU under the grant agreement n.691218 and by the EU FP7 project eVACUATE under the grant agreement n.313161.

Copyright \copyright 2017 IEEE. Personal use of this material is permitted. However, permission to use this material for any other purposes must be obtained from the IEEE by sending an email to pubs-permissions@ieee.org.}

}

\maketitle

\begin{abstract}
Detection of moving objects in videos is a crucial step towards successful surveillance and monitoring applications. A key component for such tasks is called background subtraction and tries to extract regions of interest from the image background for further processing or action. For this reason, its accuracy and real-time performance is of great significance. Although, effective background subtraction methods have been proposed, only a few of them take into consideration the special characteristics of thermal imagery. In this work, we propose a background subtraction scheme, which models the thermal responses of each pixel as a mixture of Gaussians with unknown number of components. Following a Bayesian approach, our method automatically estimates the mixture structure, while simultaneously it avoids over/under fitting. The pixel density estimate is followed by an efficient and highly accurate updating mechanism, which permits our system to be automatically adapted to dynamically changing operation conditions. We propose a reference implementation of our method in reconfigurable hardware achieving both adequate performance and low power consumption. Adopting a High Level Synthesis design, demanding floating point arithmetic operations are mapped in reconfigurable hardware; demonstrating fast-prototyping and on-field customization at the same time.
\end{abstract}


%
\IEEEpeerreviewmaketitle

\section{Introduction}
\label{sec:intriduction}
\IEEEPARstart{H}{igh} level computer vision applications, ranging from video surveillance and monitoring to intelligent vehicles, utilize visible spectrum information. However, under certain environmental conditions, this type of sensing can be severely impaired. This emerges the necessity for imaging beyond the visible spectrum, exploiting thermal sensors, 
which are equally applicable both for day and night scenarios, while at the same time are less affected by illumination changes. 

However, thermal sensors present their own unique challenges. First, they have low signal-to-noise ratio (SNR) implying the existence of noisy data.  Second, there is a lack of color and texture information deteriorating visual content interpretation \cite{wang_improved_2010}. Third, objects are not thermally homogeneous and are exposed to variations of sun illumination 
\cite{davis_background-subtraction_2007}. All these issues complicate pixel modeling especially when applying for object categorization and foreground/background detection.

For many high level applications, either they use visual-optical videos \cite{porikli_achieving_2006, tuzel_pedestrian_2008} 
or thermal data 
\cite{wang_improved_2010, jungling_feature_2009, yadav2016combined, sharma2016fisher}, the task of background subtraction constitutes a key component for locating moving objects \cite{bouwmans2014background}. The most common approach to model the background is to use mixtures of Gaussians, the number of which is assumed to be a priori known. While such assumption is valid for sensors capturing the visible spectrum, mainly due to their ultra high resolution accuracy, that is, high SNR, they are inappropriate for thermal data. Selection of a large number of components results in modeling the noise and therefore it reduces discrimination performance. On the contrary, low number of components yields approximate modeling that fails to capture the complexity of a scene. Consequently, methods that automatically estimate the most suitable components to fit the statistics of thermal data are important towards an efficient background subtraction scheme. 

Furthermore, to increase the penetration of thermal sensors to the surveillance market, embedded 
acceleration methods are needed. This means that the background subtraction algorithms should be re-designed to be implemented in low power devices. This way, the benefits are twofold. First, we offload a significant computation load near the source of the data, 
and thus, bandwidth is saved as only the region of interest (or an event) is transmitted. Second, costly operations are executed in low-power embedded hardware saving valuable resources.

\subsection{Related Work} 
\label{ssec:related work}

Background subtraction techniques applied on visual-optical videos model the color properties of objects \cite{herrero_background_2009} 
and can be classified into three main categories \cite{el_baf_fuzzy_2009}: {\it basic background modeling}, 
{\it statistical background modeling} 
and {\it background estimation}. 
The most used methods are the statistical ones due to their robustness to critical application scenarios. 

One common approach for the statistical modeling of visual content is based on the exploitation of Gaussian Mixture Models (GMMs). In the work of \cite{greenspan2004probabilistic} GMMs are utilized to create space-time representations in video segment level. However, the specific task of background subtraction requires the estimation of a pixel level background model. 

Towards this direction, the work of Stauffer and Grimson \cite{stauffer_adaptive_1999} is one of the best known approaches. It uses a GMM with a fixed number of components to estimate per-pixel density. The work of \cite{makantasis_student-t_2012} proposes a Student-t mixture model improving compactness and robustness to noise and outliers. The work of \cite{haque2008stable} proposes a background subtraction algorithm based on GMMs with only one user-tunable parameter. In \cite{lee2005effective} a GMM-based background modeling that incorporates incremental EM type of learning is proposed in order to improve convergence speed and stability of learnt models. These works assume that the number of components is a priori known. 
Following this assumption, the intensities of all pixels are represented using GMMs, all of the same number of components. However, in many real world applications this assumption can be very restrictive, because the intensities of different pixels may need different number of components for being accurately modeled.

The works of \cite{zivkovic_improved_2004} and \cite{zivkovic_efficient_2006} extend the method of \cite{stauffer_adaptive_1999} by introducing a user defined threshold to estimate the number of components. However, this rule is application dependent and not directly derived from the data. 
Another extension of \cite{stauffer_adaptive_1999} is proposed in \cite{chan2011generalized}, where a patch-wise background model is  estimated using dynamic texture. However, patch level decisions may produce very coarse results when low resolution videos, such as the thermal ones, need to be processed.

An alternative approach that makes no a priori assumptions on the number of components is presented in the work of \cite{han2004sequential}. This work proposes a recursive density approximation method that relies on the propagation of density modes, which are detected by using the mean shift. Although, mean shift is a great nonparametric technique, it is computationally intensive and thus not suitable for low power hardware implementations.

The work of \cite{haines_background_2014} proposes the exploitation of a Dirichlet Process Mixture Model (DPMM). This method automatically estimates the number of components by utilizing sampling techniques. However, sampling algorithms are computational intensive and memory inefficient and thus inappropriate for real-time applications, as the ones we focus on. To address this problem, the authors of \cite{haines_background_2014} suggest a GPU implementation. 

Another approach for modeling data distributions using GMMs is presented in \cite{priebe1994adaptive}. This work focuses on the general problem of density estimation. Based on the property of GMMs to approximate arbitrarily close any distribution, it estimates the true distribution by creating a sufficiently large number of components, which may be very "close". For density estimation problem, creating such components is not an issue since it may increase approximation accuracy. However, when one needs to design and develop an algorithm for low power hardware devices, as in our case, then this algorithm should keep in memory as few as possible parameters.

Finally, when someone knows some important features of foreground and/or background objects, supervised learning techniques can be utilized. For example in the work of \cite{ravichandran2012long} a supervised learning approach is proposed for discriminating fire than the background. However, in the general problem of background subtraction it is very uncommon to known some specific features of foreground and/or background in advance. 

Techniques that use visual/optical data present the drawback that objects' properties are highly affected by scene illumination, making the same object to look completely different under different lighting or weather conditions. Although, thermal imagery can provide a challenging alternative for addressing this difficulty, there exist few works for thermal data. 

The authors of \cite{davis_fusion-based_2005, davis_background-subtraction_2007} exploit contour saliency and a unimodal background modeling technique to extract foreground objects. However, unimodal models are not usually capable of capturing the background dynamics and its complexity. Baf \textit{et al.} in \cite{el_baf_fuzzy_2009} present a fuzzy statistical method for background subtraction to incorporate uncertainty into the GMM. Elguebaly and Bouguila in \cite{elguebaly_finite_2013} propose a finite asymmetric generalized Gaussian mixture model for object detection. However, both of these methods require a predefined maximum number of components, presenting therefore limitations when they are applied on uncontrolled environments. 

Dai \textit{et al.} in \cite{dai_pedestrian_2007} propose a method for pedestrian detection and tracking using thermal imagery. This method consists of a background subtraction technique that exploits a two-layer representation (foreground/background) of frame sequences. However, they assume that the foreground is restricted to moving objects, a consideration which is not sufficient for dynamically changing environments. One way to handle the aforementioned difficulties is to introduce a background model, the parameters and the structure of which are directly estimated from the data, while at the same time it takes into account the specific characteristics of thermal imagery.

The computational cost, and thus the performance, of a background subtraction algorithm is always an issue as it usually performs poor in CPUs. One of the first attempts for real-time performance
was the work of \cite{wren_pfinder:_1997} implemented in SGI O2 computer. Since then, many implementations in GPUs were proposed. In \cite{carr2008gpu} and \cite{pham2010gpu} implementations based on the model of \cite{stauffer_adaptive_1999} achieving real-time performance even for High-Definition (HD) resolutions are presented. In the work of \cite{zhang2014gpu} the authors managed to accelerate the algorithm of \cite{zivkovic_improved_2004} up to 1080p resolution of 60fps. However, GPUs cannot be considered low power devices, which can be seen as a disadvantage especially for 24/7 surveillance systems. 

This gap is addressed from Field Programmable Gate Arrays (FPGA) accelerators. In the work of \cite{kristensen2008embedded} and \cite{jiang2009hardware}, a real-time video surveillance system using a GMM, that also handles memory bandwidth reduction requirements, is proposed. Other approaches such as the work of \cite{genovese2013fpga} propose accelerators of the GMM algorithm in reconfigurable hardware reaching 24fps for HD video. The same authors claim even better performance of 91fps in HD video in their improved work of \cite{genovese2014asic} if a Xilinx Virtex 4 device is used. However, the main limitation to achieve this performance is the memory bandwidth which becomes the main bottleneck in the pipeline. The requested bandwidth for this performance is about 8GB/sec where FPGA boards usually hold 64bit-wide Dynamic Random Access Memory (DRAM) clocked in a range of 100-200 MHz. As a result the memory subsystem can support at least one order of magnitude lower bandwidth. This means that we need technologies for reducing memory requirements in case that the background subtraction algorithm is adapted to be implemented under reconfigurable hardware architectures. 

\subsection{Our Contribution}
\label{sec:our contribution}
The main contribution of this paper is the design of a background subtraction system (as a whole) that is completely data driven, takes into account the specific characteristics of thermal imagery and is suitable for implementation in low power and low memory hardware.

This work extends our previous works in \cite{makantasis2015variational, nikitakis2016novel}. Our method exploits GMMs with unknown number of components, which are dynamically estimated directly from the data. 
In particular, the Variational Inference framework is adopted to associate the functional structure of the model with real data obtained from thermal images. Variational Inference belongs to the class of probability approximation methods, which try to estimate the approximate posterior distribution by minimizing the KL-divergence between the approximate and the true posteriors. As it has been shown in \cite{bernardo2003variational} and \cite{teschendorff2005variational}, when the number of samples tends to infinity the lower variational bound approaches the BIC criterion for model selection. When someone targets low power hardware devices, and thus must keep in memory as few as possible parameters, this feature is very important, since through Variational Inference the true distribution is approximated without over/under fitting (creates the ''right'' number of components). 

The adopted approach, instead of treating the mixing coefficients of the components as single parameters, it considers them as probability distributions. Under this framework, we need to estimate forms of probability distributions that best fit data properties, instead of fitting an a priori known number of components to the captured data. Then, the Expectation-Maximization (EM) algorithm is adopted to estimate model parameters. To compensate computational challenges 
we utilize conjugate priors for deriving analytical equations for model estimation. 
Updating procedures are incorporated to allow dynamic model adaptation. Our updating method avoids the use of accumulated data from previous time instances, resulting in low memory requirements. Such a scheme assists the implementation of an in-camera module suitable for devices of low power and memory demands. 

This paper is organized as follows: Section {\ref {sec:variational inference for gaussian mixture modeling}} introduces the Variational Inference framework, 
while Section {\ref {sec:derivation of random variables distribution}} describes the algorithm for optimally estimating the model parameters. In Section {\ref {sec:random variables optimization}}, we present the EM optimization that best fits model parameters to the data. 
A threshold independent on-line updating algorithm is introduced in Section {\ref {Sec: updating}}. 
The in-camera reconfigurable architecture is discussed in Section {\ref {sec:in-camera acceleration architecture}}, while experimental results are presented in Section {\ref {sec:experimental validation}}.q Finally, Section {\ref {sec: conclusions}} draws the conclusions of the paper.

\section{Variational Inference for Gaussian Mixture Modeling}
\label{sec:variational inference for gaussian mixture modeling}

\subsection{Gaussian Mixture Model Fundamentals}
\label{ssec:gaussian mixture model fundamentals}

The Gaussian mixture distribution has the following form;
\begin{equation}
p(x|\bm \varpi, \bm \mu, \bm \tau) = \sum_{\bm z} p(\bm z | \bm \varpi) p(x| \bm z, \bm \mu, \bm \tau),
\label{eq:gaussian_mm_z}
\end{equation}
where $p(\bm z|\bm \varpi)$ and and $p(x|\bm z, \bm \mu, \bm \tau)$ are in the form of
\begin{equation}
p(\bm z | \bm \varpi) = \prod_{k=1}^{K} \varpi_k^{z_k}, 
\label{eq:p_z}
\end{equation}
\begin{equation}
p(x|\bm z, \bm \mu, \bm \tau) = \prod_{k=1}^{K} \mathcal{N}(x|\mu_k, \tau_k^{-1})^{z_k}.
\label{eq:p_x}
\end{equation}
In Eq.(\ref{eq:p_z}) and Eq.(\ref{eq:p_x}), $\mathcal N(\cdot)$ represents the Gaussian distribution, $K$ is the number of components, variables $\{\varpi_k\}_{k=1}^K$ refer to the mixing coefficients that represent the proportion of data that belong to each component and which satisfy $0 \leq \varpi_k \leq 1$ and $\sum_{k=1}^K \varpi_k = 1$. Variable $x$ corresponds to the intensity of a pixel (i.e., the observed variable) and $\{\mu_k\}_{k=1}^K$, $\{\tau_k\}_{k=1}^K$ stand for the mean values and precisions of the Gaussian components respectively. The $K$-dimensional vector $\bm z=[z_1, ..., z_K]$ is a binary latent variable in which a particular element is equal to one and all other elements are equal to zero, such as $\sum_{k=1}^K z_k= 1$ and $p(z_k=1) = \varpi_k$. This vector is related to the number of components that are used for modeling pixels intensities. In the work of \cite{stauffer_adaptive_1999} the value of $K$ is assumed to be a priori known, while in our case, this value is estimated directly from the data.

Eq.(\ref{eq:gaussian_mm_z}) models the effect of one sample. Given a set $\bm X = \{x_1,...,x_N\}$ of $N$ pixel intensities (i.e., observed data), we conclude to a set of $N$ latent variables, $\bm Z = \{\bm z_1,...,\bm z_N \}$. Each $\bm z_n$ is a $K$-dimensional binary vector with one element equals one and all the others equal zero, such as $\sum_{k=1}^K z_{nk}= 1$. Then, Eq.(\ref{eq:p_z}) and Eq.t(\ref{eq:p_x}) are transformed to
\begin{equation}
p(\bm Z | \bm \varpi) = \prod_{n=1}^{N} \prod_{k=1}^{K} \varpi_k^{z_{nk}},
\label{eq:p_Z}
\end{equation}
\begin{equation}
p(\bm X|\bm Z, \bm \mu, \bm \tau) = \prod_{n=1}^{N} \prod_{k=1}^{K} \mathcal{N}(x_n|\mu_k, \tau_k^{-1})^{z_{nk}}.
\label{eq:p_X}
\end{equation}
The goal is to estimate a background model exploiting pixel intensities, that is, to calculate the values of $\bm \varpi$, $\bm \mu$ and $\bm \tau$, involved in the probability $p(x|\bm \varpi, \bm \mu, \bm \tau)$.

\subsection{Distribution Approximation through Variational Inference}
\label{ssec:distribution approximation through variational inference}

 In case that variable $K$ of a GMM is a priori known, the values of $\bm \varpi$, $\bm \mu$ and $\bm \tau$ can be straightforward calculated using the methods of \cite{stauffer_adaptive_1999,zivkovic_improved_2004}, which exploit the k-means algorithm. For many real-life application scenarios, as the one this paper targets, it is better to let variable $K$ fit the statistics of the data  (i.e., let variable $K$ be unknown). In such cases, one way to estimate $K$ is to apply computationally expensive methods through sampling algorithms or to build several models of different number of components and then select the best one. An alternative computationally efficient approach, adopted in this paper, is to exploit the Variational Inference framework. More specifically, instead of treating the mixing coefficients $\bm \varpi$ as single parameters, which requires the knowledge of $K$, we treat them as probability distributions. This way, we are able to estimate the coefficients $\bm \varpi$ independently from $K$. Such an approach keeps computational complexity low since it avoids sampling methods or experimentation on different number of components. 

Let us denote as $\bm Y=\{\bm Z, \bm \varpi, \bm \mu, \bm \tau\}$ a set which contains model parameters and the respective latent variables. Let us also denote as  $q(\bm Y)$ the variational distribution of $\bm Y$. Our objective is to estimate $q(\bm Y)$ to be as close as possible to $p(\bm Y | \bm X)$ for a given observation $\bm X$. Regarding similarity between two distributions, the Kullback-Leibler divergence, 
\begin{equation}
\label{eq:kullback-leibler}
KL(q||p) = \int q(\bm Y) \ln \frac{q(\bm Y)} {p(\bm Y | \bm X)} d\bm Y,
\end{equation}
 is used. $KL(q||p)$ has to be minimized since it is a non negative quantity, which equals zero only if $q(\bm Y)=p(\bm Y | \bm X)$. 

In the context of the most common type of Variational Inference, known as \textit{mean-field approximation}, the variational distribution is assumed to be factorized over $M$ disjoint sets such as $q(\bm Y) = \prod_{i=1}^{M}q_i(\bm Y_i)$. Then, as shown in \cite{bishop_pattern_2007}, the optimal solution $q_j^*(Y_j)$ that minimizes $KL(q||p)$ metric is given by
\begin{equation}
\ln q_j^*(\bm Y_j) = \mathbb{E}_{i \neq j} [ \ln p(\bm X, \bm Y) ] + \mathcal{C},
\label{eq:q_star}
\end{equation}
where $\mathbb{E}_{i \neq j} [ \ln p(\bm X, \bm Y) ]$ is the expectation of the logarithm of the joint distribution over all variables that do not belong to the $j^{th}$ partition and $\mathcal{C}$ is a constant. Eq.(\ref{eq:q_star}) indicates the presence of circular dependencies between the variables that belong to different partitions. Thus, estimating the optimal distribution over all variables suggests the exploitation of an iterative process such as the EM algorithm (see Section \ref{sec:random variables optimization}).

\section{Optimal distributions over model parameters}
\label{sec:derivation of random variables distribution}

In this section, we present the analytical form for the optimal distributions $q_j^*(Y_j)$, considering the model coefficients and the latent variables; i.e., $q_Z^*(\bm Z)$, $q_{\varpi}^*(\bm \varpi)$, $q_{\tau}^*(\bm \tau)$ and $q_{\mu|\tau}^*(\bm \mu|\bm \tau)$. For simplifying the notation, in the following the superscript of optimal distributions and the subscript for the $j^{th}$ partition are omitted.

\subsection{Factorized Form of the Joint Distribution}
To estimate $q(\bm Y)$, we require to rewrite the right hand of Eq.(\ref{eq:q_star}), that is, the joint distribution $p(\bm X, \bm Y)$, as a product 
\begin{equation}
p(\bm X, \bm Y) = p(\bm X|\bm Z, \bm \mu, \bm \tau) p(\bm Z|\bm \varpi) p(\bm \varpi) p(\bm \mu,\bm \tau).
\label{eq:joint_factorization}
\end{equation} 	
The distributions $p(\bm X|\bm Z, \bm \mu, \bm \tau)$ and $p(\bm Z|\bm \varpi)$ are already known from Eq.(\ref{eq:p_X}) and Eq.(\ref{eq:p_Z}). Thus, we need to define the prior distribution $p(\bm \varpi)$ and the joint distribution $p(\bm \mu , \bm \tau)$. In this paper, conjugate priors 
are adopted to estimate the distributions $p(\bm \varpi)$ and $p(\bm \mu , \bm \tau)$. Such an approach is computational efficient since it avoids implementation of the expensive sampling methods yielding analytical solutions. 

We start our analysis by the prior distribution $p(\bm \varpi)$. In particular, since $p(\bm Z | \bm \varpi)$ has the form of a multinomial distribution, [see Eq.(\ref{eq:p_Z})], its conjugate prior is given by 
\begin{equation}
p(\bm \varpi) = \frac{\Gamma(K\lambda_0)}{\Gamma(\lambda_0)^K} \prod_{k=1}^{K}\varpi_k^{\lambda_0-1}.
\label{eq:varpi_prior}
\end{equation} 
Eq.(\ref{eq:varpi_prior}) is a Dirichlet distribution 
with $\Gamma(\cdot)$ stands for the Gamma function and scalar $\lambda_0$ a control parameter. The smaller the value of  $\lambda_0$ is, the larger the influence of the data rather than the prior on the posterior distribution $p(\bm Z|\bm \varpi)$. The choice of setting the parameter $\lambda_0$ as a scalar instead of a vector of different values for each mixing coefficient, is due to the fact that we adopt an uninformative prior framework, that is not preferring a specific component against the others.  

Similarly, $p(\bm \mu, \bm \tau)$ is the prior of $p(\bm X|\bm Z, \bm \mu, \bm \tau)$ which is modeled through Eq. (\ref{eq:p_X}). The conjugate prior of (\ref{eq:p_X}) takes the form of a Gaussian-Gamma distribution 
since both $\bm \mu$ and $\bm \tau$ are unknown. Subsequently, the joint distribution $p(\bm \mu, \bm \tau)$ can be modeled as 
\begin{subequations}
\begin{align}
p(\bm \mu, \bm \tau & )  = p(\bm \mu | \bm \tau)p(\bm \tau) \\
& = \prod_{k=1}^{K} \mathcal{N}(\mu_k|m_0,(\beta_0\tau_k)^{-1})Gam(\tau_k|a_0,b_0),
\end{align}
\label{eq:joint_mu_tau_prior}
\end{subequations}
where $Gam(\cdot)$ denotes the Gamma distribution. Again, an uninformative prior framework is adopted meaning that no specific preference about the form of the Gaussian components is given. The parameters $m_0$, $\beta_0$, $a_0$ and $b_0$  are discussed in Section \ref{priors}.
In the following, the forms of optimal variational distributions are presented using the results from Appendix \ref{ap:appendix}.

\subsection{Optimal $q^*(\bm Z)$ Distribution}
Using Eq.(\ref{eq:q_star}) and the factorized form of Eq.(\ref{eq:joint_factorization}), the distribution of the optimized factor $q^*(\bm Z)$ is given by a Multinomial distribution of the form
\begin{subequations}
\begin{align}
& q^*(\bm Z) = \prod_{n=1}^{N}\prod_{k=1}^{K}\bigg(\frac{\rho_{nk}}{\sum_{j=1}^{K}\rho_{nj}}\bigg)^{z_{nk}} = \\ 
& \:\:\:\:\:\:\:\:\:\:\:\:= \prod_{n=1}^{N}\prod_{k=1}^{K} r_{nk}^{z_{nk}},
\label{eq:q_Z_optimized_dirichlet_b}
\end{align}
\label{eq:q_Z_optimized_dirichlet}
\end{subequations}  
where quantity $\rho_{nk}$ is given as 
\begin{equation}
\begin{aligned}
\rho_{nk} = \exp\bigg(& \mathbb{E}\big[\ln \varpi_k \big] + \frac{1}{2}\mathbb{E}\big[\ln \tau_k\big] -\frac{1}{2}\ln2\pi - \\ &-\frac{1}{2} \mathbb{E}_{\bm \mu, \bm \tau}\big[(x_n-\mu_k)^2\tau_k \big]\bigg).
\end{aligned}
\label{eq:pho_nk}
\end{equation}
Due to the fact that $q^*(\bm Z)$ is a Multinomial distribution we have that its expected value $\mathbb{E}[z_{nk}]$ will be equal to $r_{nk} $

\subsection{Optimal $q^*(\bm \varpi)$ Distribution}
\label{opt}
Using Eq.(\ref{eq:joint_factorization}) and Eq.(\ref{eq:q_star}) the variational distribution of the optimized factor $q^*(\bm \varpi)$ is given by the Dirichlet distribution
\begin{equation}
q^*(\bm \varpi) = \frac{\Gamma(\sum_{i=1}^{K}\lambda_i)}{\prod_{j=1}^{K}\Gamma(\lambda_j)} \prod_{k=1}^{K}\varpi_k^{\lambda_k - 1}.
\label{eq:q_star_varpi_derivation}
\end{equation}
Variable $\lambda_k$ is equal to $N_k + \lambda_0$, while $N_k=\sum_{n=1}^{N}r_{nk}$ represents the proportion of data that belong to the $k$-th component.

\subsection{Optimal $q^*(\mu_k | \tau_k)$ distribution}
Similarly, the variational distribution of the optimized factor $q^*(\mu_k, \tau_k)$ is given by a Gaussian distribution of the form 
\begin{equation}
q^*(\mu_k|\tau_k) = \mathcal{N}(\mu_k|m_k, (\beta_k \tau_k)^{-1}),
\label{eq:q_star_mu_distribution}
\end{equation}
where the parameters $m_k$ and $\beta_k$ are given by 
\begin{subequations}
\begin{align}
\beta_k & = \beta_0 + N_k, \\
m_k & = \frac{1}{\beta_k}\Big(\beta_0 m_0 + N_k \bar x_k\Big).
\end{align}
\label{eq:q_star_mk_tauk}
\end{subequations}
Variable $\bar x_k$ is equal to $\frac{1}{N_k}\sum_{n=1}^{N}r_{nk}x_n$ and represents the centroid of the data that belong to the $k$-th component.

\subsection{Optimal $q^*(\tau_k)$ distribution}

After the estimation of $q^*(\mu_k|\tau_k)$, the variational distribution of the optimized factor $q^*(\tau_k)$ is given by a Gamma distribution of the following form 
\begin{equation}
q^*(\tau_k) = Gam(\tau_k|a_k, b_k),
\label{eq:q_star_tau_distribution}
\end{equation}
while the parameters $a_k$ and $b_k$ are given as  
\begin{subequations}
\begin{align}
a_k & = a_0 +  \frac{N_k}{2} 
\label{eq:q_star_tauk_a},\\
b_k & = b_0 + \frac{1}{2}\bigg(N_k\sigma_k + \frac{\beta_0 N_k}{\beta_0 + N_k}\big(\bar x_k - m_0\big)^2 \bigg),
\label{eq:q_star_tauk_b}
\end{align}
\label{eq:q_star_tauk}
\end{subequations}
where $\sigma_k = \frac{1}{N_k}\sum_{n=1}^{N}(x_n-\bar x_k)^2$.

\section{Distribution Parameters optimization}
\label{sec:random variables optimization}

In Section \ref{sec:derivation of random variables distribution}, we derive approximations of the random variable distributions. While the works of \cite{stauffer_adaptive_1999} and \cite{zivkovic_improved_2004} adopt k-means algorithm to approximately estimate the parameters of the background model, in this work, the EM algorithm is employed to optimally estimate the coefficient distributions that best fit the observations. 

\subsection{The EM Optimization Framework}
\label{EM}

{\bf E-Step:} Let us assume the $t$-th iteration of the EM optimization algorithm. Then, during the E-step, only the value of $r_{nk}$ is readjusted according to the statistics of the currently available observed data. Variable $r_{nk}$ actually expresses the degree of fitness of the $n$-th datum to the $k$-th cluster, as derived from Eq.(\ref{eq:q_Z_optimized_dirichlet}). Due to the fact that $q^*(\bm \varpi)$ is a Dirichlet distribution and $q^*(\tau_k)$ is a Gamma distribution, the following holds 

\begin{subequations}
\begin{align}
& \ln \tilde{\tau_k}(t) \equiv \mathbb{E}\big[\ln \tau_k(t)\big] = \Psi(a_k(t)) - \ln b_k(t), \\
& \ln \tilde{\varpi_k}(t) \equiv \mathbb{E}\big[\ln \varpi_k(t)\big] = \Psi(\lambda_k(t)) - \Psi\bigg(\sum_{k=1}^{K} \lambda_k(t)\bigg), \\
& \mathbb{E}\big[\tau_k(t)]\big] = \frac{a_k(t)}{b_k(t)},
\end{align}
\label{eq:tilde_pi_tilde_tau}
\end{subequations}
where $\Psi(\cdot)$ is the digamma function. In Eq.(\ref{eq:tilde_pi_tilde_tau}), we set $\ln \tilde{\tau_k}(t) \equiv \mathbb{E}\big[\ln \tau_k(t)\big]$ and $\ln \tilde{\varpi_k}(t) \equiv \mathbb{E}\big[\ln \varpi_k(t)\big]$ to simplify the notation of the following equations. Then, 
\begin{equation}
\begin{aligned}
r_{nk}(t+1) & \propto \tilde{\varpi_k}(t) \tilde{\tau_k}(t)^{1/2} \\ & \exp \bigg(-\frac{a_k(t)}{2b_k(t)}\big(x_n-m_k(t)\big)^2 - \frac{1}{2\beta_k(t)} \bigg) \end{aligned}
\label{eq:r_nk}
\end{equation}
by substituting Eq.(\ref{eq:tilde_pi_tilde_tau}) into Eq.(\ref{eq:pho_nk}) and using Eq.(\ref{eq:q_Z_optimized_dirichlet}). In Eq.(\ref{eq:r_nk}), $r_{nk}(t+1)$ expresses the degree of fitness of the $n$-th datum to the $k$-th cluster at the next $t$+1 iteration  of the algorithm. 

{\bf M-Step:} During the M-step, we keep fixed the value of  $r_{nk}(t)$, as it has been calculated through the E-Step. Then, we update the values of the background model coefficients, which will allow us to re-estimate the degree of fitness $r_{nk}$ at the next iteration stage, exploiting Eq.(\ref{eq:r_nk}).

Particularly, initially, the variables $N_k(t+1)$ and $\lambda_k(t+1)$ are estimated, based on the statements of Section \ref{opt} and the  $r_{nk}(t+1)$ of Eq.(\ref{eq:r_nk}),
\begin{subequations}
\begin{align}
& N_k(t+1)=\sum_{n=1}^{N}r_{nk}(t+1),\\
& \lambda_k(t+1)=N_k(t+1) + \lambda_0. 
\end{align}
\label{eq:N and Lambda}
\end{subequations}
These are the only variables that are needed for updating model parameters 
using Eq.(\ref{eq:q_star_varpi_derivation}), Eq.(\ref{eq:q_star_mu_distribution}) and Eq.(\ref{eq:q_star_tau_distribution}).

The distribution $q^*(\bm \varpi(t+1))$ of the model coefficients is computed based on Eq.(\ref{eq:q_star_varpi_derivation}). The value  $\lambda_0$ is given in Section \ref{priors}. We recall that in our approach, the number of components that the background content is composed to is not a priori known. For this reason, the mixing coefficients of the background model are treated as probability distributions and not as single parameters. Due to this fact, we can initialize the EM algorithm by setting the number of components to be smaller than or equal to a maximum value, coinciding with the number of observed data, that is, $K_{max}\leq N$. Then, the probability coefficients distributions re-arrange the number of components, in order to best fit the statistics of the observations. This is achieved through EM optimization.

In the following, the parameters $a_k(t+1)$ and $b_k(t+1)$ are updated to define the Gamma distribution of $q^*(\tau_k(t+1))$ that best fit the observations through Eq.(\ref{eq:q_star_tauk}). Again, the priors $a_0$, $b_0$ and $\beta_0$ are given in Section \ref{priors}. 

Next, the distribution $q^*(\mu_k(t+1)| \tau_k(t+1))$ is updated exploiting $\tau_k(t+1)$. In order to do this, we need to update $\beta_k(t+1)$ and    $m_k(t+1)$ based on Eq.(\ref{eq:q_star_mk_tauk}). 
The E and M steps are repeated sequentially until the values for model parameters are not significantly changing. As shown in \cite{boyd_convex_2004} convergence of EM algorithm is guaranteed because bound is convex with respect to each of the factors $q(\bm Z)$, $q(\bm \varpi)$, $q(\bm \mu|\bm \tau)$ and $q(\bm \tau)$. 

During model training the mixing coefficient for some of the components takes value very close to zero. Components with mixing coefficient less than $1/N$ are removed (we require each component to model at least one observed sample) and thus after training, the model has automatically determined the right number of Gaussian components.
 
\subsection{Initialization Aspects}
\label{model}

The k-means++ algorithm \cite{arthur2007k} is exploited to initialize the EM algorithm at $t=0$. The k-means++ presents advantages compared to the conventional k-means, since it is less depended on initialization. It has to be mentioned that fuzzy versions of k-means are not suitable for the initialization process, since each sample should belong to exactly on cluster/component. The k-means++ creates an initial partition of the data used to initialize EM algorithm. Then, at the updating stages of the algorithm (Section \ref{EM}), the probabilities of each observed datum to belong to one of the $K_{max}$ available clusters, expressed through $r_{nk}$, are updated. This way, the final number of clusters are dynamically refined according to the statistical distributions of the image pixel intensities.

Let us denote as $\hat N_k=N_k(t=0)$ the number of observations that belong to $k$-th cluster at iteration $t=0$. Then, an initial estimate of the mixing coefficients is $\varpi_k(t=0) = \hat N_k/N$, meaning that the significance of the $k$-th component is proportional to the number of data that belong to the $k$-th cluster. Thus, the initialization of $\lambda_k(t=0)=N\varpi_k(t=0)+\lambda_0$, [see Eq.(\ref{eq:q_star_varpi_derivation})] expresses the number of observations associated with each component.
The parameters $a_k(t=0)$, $b_k(t=0)$, $\beta_k(t=0)$ and $m_k(t=0)$ are initially estimated from Eq.(\ref{eq:q_star_tauk}) and Eq.(\ref{eq:q_star_mk_tauk}), considering the knowledge of the priors parameters as discused in Section \ref{priors}. Finally, the model parameter $\tau_k(t=0)$ is given as inverse proportional of the variance of the data of the $k$-th initial cluster, that is, $\tau_k(t=0)=\hat v_k^{-1}(t=0)$.

\subsection{Priors Parameters}
\label{priors}
The parameter $\lambda_0$ in Eq.(\ref{eq:varpi_prior}) can be interpreted as the effective prior number of observations associated with each component. However, we do not have any prior information regarding this number. In order to use an uninformative prior and maximize the influence of the data over the posterior distribution we set $\lambda_0=1$, see \cite{yang1996catalog}.

Relations Eq.(\ref{eq:q_star_tauk}) and Eq. (\ref{eq:q_star_tauk_b}) suggest that the values of parameters $a_k$ and $b_k$ are primarily affected by the data and not by the prior, when the values of the parameters $a_0$ and $b_0$ are close to zero. For this reason we set $a_0$ and $b_0$ to a very small value ($10^{-3}$ in our implementation).

Similarly, we initialize $m_0$ as the mean value of the observed data and precision $\beta_0=\frac{b_0}{a_0 v_0}$, where $v_0$ is the variance of the observed data. We use uninformative priors, since we do not have any information regarding neither the number of components nor their true mean and variance values.

\section{Online Updating Mechanism and Background Subtraction}
\label{Sec: updating}
Using the aforementioned approach, we fit a model to the background considering a pool of $N$ observed data. In this section, an adaptive strategy that is threshold-independent and memory efficient is presented. Such an approach permits implementation of the proposed algorithm to an in-camera embedded hardware of low power and memory requirements. This way we deliver thermal sensors embedding with the capability of detecting moving objects in real-time. Furthermore, by exploiting the updating mechanism the presented system can online process streams of frames yielding a small computational time. So, in a sense, it can handle big data volumes.

Let us denote as $x_{new}$ a new observed sample. Then, a decision is made whether $x_{new}$ can be approximated by our best fitted model or not. For this reason, the best matched Gaussian component $c$  to $x_{new}$ is estimated by minimizing the Mahalanobis distance $D_k$,
\begin{equation}
c = \arg \min_k D_k = \arg \min_k \sqrt{(x_{new}-\mu_k)^2 \tau_k},
\label{eq:Mahalanobis}
\end{equation}
where $\mu_k$ and $\tau_k$ stand for the mean and precision of the $k$-th component. We use Mahalanobis distance, since this is a reliable distance measure between a point and a distribution. Then, $x_{new}$ belongs to $c$ with probability
\begin{equation}
p(x_{new}|\mu_c, \tau_c) = \mathcal{N}(x_{new}|\mu_c, \tau_c^{-1}).
\label{eq:p_mix}
\end{equation}

\subsection{Threshold Independent} 
Conventionally, Eq.(\ref{eq:p_mix}) implies a threshold to determine the probability limit over which the new sample $x_{new}$ belongs to $c$. To overcome threshold limitations, the following adaptive approach is adopted in this paper. 

Let us denote as $\Omega$ the image pixel responses over a fixed time span. Then, we model the probability to observe the new sample $x_{new}$ in a region of range $2\epsilon$ centered at $x_{new}$ as  

\begin{equation}
p(x_{new};\epsilon) = \frac{N_{\epsilon}}{N} \mathcal{U}(x_{new};x_{new}-\epsilon, x_{new}+\epsilon),
\label{eq:p_xnew}
\end{equation}
where $N_{\epsilon}=\big|\{ x_i \in \Omega : x_{new}-\epsilon \leq x_i \leq x_{new}+\epsilon \}\big|$ is the cardinality of the set that contains samples $\epsilon$-close to $x_{new}$ and $\mathcal{U}(x_{new};x_{new}-\epsilon, x_{new}+\epsilon)$ is a Uniform distribution with lower and upper bounds that equal to $x_{new}-\epsilon$ and $x_{new}+\epsilon$. respectively. 

Eq.(\ref{eq:p_xnew}) suggests that the probability to observe the $x_{new}$ is related to the portion of data that have been already observed around $x_{new}$. By increasing the neighborhood around $x_{new}$, i.e., increasing the value of $\epsilon$, the quantity $\mathcal{U}(x_{new};x_{new}-\epsilon, x_{new}+\epsilon)$ is decreasing, while the value of $N_{\epsilon}$ is increasing. Therefore, we can estimate the optimal range $\epsilon^*$  around $x_{new}$ that maximizes Eq. (\ref{eq:p_xnew})
\begin{equation}
\epsilon^* = \arg \max_{\epsilon} p(x_{new};\epsilon).
\label{eq:epsilon}
\end{equation}

Based on the probabilities $p(x_{new}|\mu_c, \tau_c)$ and $p(x_{new};\epsilon^*)$, which are exclusively derived by the observations, we can define our decision making mechanism. Concretely, if
\begin{equation}
p(x_{new}|\mu_c, \tau_c) \geq p(x_{new}|\epsilon^*), 
\label{eq:decision}
\end{equation}
the new observed sample $x_{new}$ can sufficiently represented by our model, i.e., the value of the new observed sample is sufficiently close to an existing Gaussian component. Otherwise, a new Gaussian component should be created, since the value of $x_{new}$ will not be close to what the model has already learnt.

\subsection{Model Updating} 
When the value of the new observed sample is sufficiently close to an existing Gaussian component, the parameters of the mixture are being updated using the \textit{following the leader} \cite{dasgupta_-line_2007} approach described as
\begin{subequations}
\begin{align}
& \varpi_k \leftarrow \varpi_k + \frac{1}{N} \big(o_k - \varpi_k\big), \\ 
& \mu_k \leftarrow \mu_k + o_k\bigg( \frac{x_{new}-\mu_k}{\varpi_k N + 1} \bigg), \\
& \sigma_k^2 \leftarrow \sigma_k^2 + o_k \bigg(\frac{\varpi_k N (x_{new}-\mu_k)^2}{(\varpi_k N + 1)^2} -\frac{\sigma_k^2}{\varpi_k N + 1}  \bigg),
\end{align}
\label{eq:updating_equations}
\end{subequations}
where $\sigma_k^2=\tau_k^{-1}$ is the variance of the $k$-th component. The binary variable $o_k$ takes value one when $k=c$ and zero otherwise.

When the new observed sample cannot be modeled by any existing component, i.e. the value of the new sample will not be close to what the model has already learnt [see Eq.(\ref{eq:decision})], a new component is created with mixing coefficient $\varpi_{new}$, mean value $\mu_{new}$ and standard deviation $\sigma_{new}$, defined as

\begin{subequations}
\begin{align}
& \varpi_{new} = \frac{1}{N},\\
& \mu_{new} = x_{new},\\
& \sigma_{new}^2 = \frac{(2\epsilon)^2 - 1}{12}.
\end{align}
\label{eq:new_component}
\end{subequations} 

Variable $\sigma_{new}^2 $ is estimated using the variance of the Uniform distribution. From Eq.(\ref{eq:new_component}), we derive that $\varpi_{new}=1/N$ since it models only one sample (the new observed one), its mean value equals the value of the new sample and its variance the variance of the Uniform distribution, whose the lower and upper bounds are $x_{new}-\epsilon$ and $x_{new}+\epsilon$ respectively. When a new component is created the values for the parameters for all the other components remain unchanged except from the mixing coefficients $\{\varpi_k\}_{k=1}^K$ which are normalized to sum $\frac{N-1}{N}$. Then, the components whose mixing coefficients are less than $\frac{1}{N}$ are removed 
and the mixing coefficients of the remaining components are re-normalized.

\subsection{Memory Efficient Implementation}
\label{ssec:online adaptation mechanism}
The main limitation of the aforementioned threshold independent approach is that it requires the storage of several observations, in order to reliably estimate the probability  $p(x_{new};\epsilon^*)$. In this section, we introduce a framework of updating the model parameters without the need of storing observations. This reduces memory requirements, and thus, it is a crucial step towards implementing our proposed system on devices of low power and memory requirements.

We recall that we have denoted as $c$ the closest component, in terms of Mahalaobis distance, to the new observed datum $x_{new}$. This component is a Gaussian distribution with mean value $\mu_c$, precision $\tau_c$ and mixing coefficient $\varpi_c$. Therefore, the quantity $N_{\epsilon}$ can be approximated as
\begin{equation}
\label{eq:N_e_approx}
N_e \approx \tilde{N}_{\epsilon} = N \varpi_c \int_{x_{new}-\epsilon}^{x_{new}+\epsilon} \mathcal{N}(t|\mu_c, \tau_c^{-1}) dt.
\end{equation}
Denoting as 
\begin{equation}
G_c(x) = \int_{-\infty}^{x} \mathcal{N}(t|\mu_c, \tau_c^{-1}) dt
\end{equation}
the cumulative Gaussian distribution of the closest Gaussian component and using Eq.(\ref{eq:N_e_approx}), $\tilde{N}_e$ is equal to 
\begin{equation}
\tilde{N}_{\epsilon} = N \varpi_c \big(G_c(x_{new}+\epsilon) - G_c(x_{new}-\epsilon)\big)
\end{equation}
Then, the probability  $p(x_{new};\epsilon)$ is approximated as
\begin{equation}
\begin{aligned}
p(x_{new};\epsilon) & \approx \tilde{p}(x_{new};\epsilon) =\\
& =\frac{\tilde{N}_{\epsilon}}{N}\mathcal{U}(x_{new};x_{new}-\epsilon, x_{new}+\epsilon).
\end{aligned}
\label{eq:p_xnew_approx}
\end{equation}

Probability $\tilde{p}(x_{new};\epsilon)$ is a continuous and unimodal function. Therefore, $\epsilon^*$ can be found either by setting the first derivative of Eq.(\ref{eq:p_xnew_approx}) equal to zero or by using a numerical approach according to which the epsilon value is increased using "sufficiently" small steps in order to approximate the point where the curvature of Eq.(\ref{eq:p_xnew_approx}) changes. This point indicates the optimal value of epsilon.  After the estimation of $\epsilon^*$, we can compute $\tilde{p}(x_{new};\epsilon^*)$. Thus,  we are able to update the mixture model by comparing $\tilde{p}(x_{new};\epsilon^*)$ to $p(x_{new}|\mu_c, \tau_c)$. 

\subsection{Background Subtraction}
\label{ssec:background subtraction}
Let us denote as $bg$ and $fg$ the classes of background and foreground pixels respectively. The aforementioned modeling process actually approximates the probability $p(x|bg)$. However, our goal is to calculate the probability  $p(bg|x)$, in order to as foreground or background a set of observations.  Hence the Bayes rule is applied;
\begin{equation}
\label{eq:bayes rule}
p(bg|x) = \frac{p(x|bg) p(bg)}{p(x|bg) + p(x|fg)}.
\end{equation}
Then, the foreground object is derived through a subtraction process. The unknown factors of Eq.(\ref{eq:bayes rule}) are $p(bg)$ and $p(x|fg)$. The probability $p(bg)$ corresponds to the prior probability of background class. In our case, we have set it to be larger than $1/2$, since the number of pixels that belong to the background class is larger than the number of pixel that belong to the foreground class. The probability $p(x|fg)$ is modeled using a uniform distribution as in \cite{haines_background_2014}. Thus, $p(x|fg)$ at arbitrary value of $x$ is $1/256$, since $x$ can take arbitrary integer values between 0 and 255. The overview of the proposed scheme is shown in Algorithm 1.

Following this approach, our method avoids outliers by assigning them to components with very low weight. This way outliers are not practically considered during background subtraction, since $p(x|bg)$ will be close to zero when $x$ is an outlier. Furthermore, by exploiting the proposed online adaptation mechanism, components assigned to outliers will be discarded after the capturing of a few new frames, because their weight will be smaller than $1/N$.

\begin{tabular}{ l }
  \hline \hline                      
  \textbf{Algorithm 1}: Overview of Background Subtraction  \\
  \hline
  1:\:\:\: capture $N$ frames \\
  2:\:\:\: create $N$-length history for each pixel \\
  3:\:\:\: initialize parameters (see Section \ref{sec:random variables optimization}) \\
  4:\:\:\: \textbf{until} convergence (training phase: Section \ref{sec:random variables optimization}) \\
  5:\:\:\:\:\:\:\:\:\: compute $r_{nk}$ using (\ref{eq:r_nk}) \\
  6:\:\:\:\:\:\:\:\:\: recompute parameters using (\ref{eq:q_star_varpi_derivation}), (\ref{eq:q_star_mk_tauk}) and (\ref{eq:q_star_tauk}) \\
  7:\:\:\: \textbf{for each} new captured frame \\
  8:\:\:\:\:\:\:\:\:\: classify each pixel as foreground or background \\
  \:\:\:\:\:\:\:\:\:\:\:\:\: (see subection \ref{ssec:background subtraction}) \\
  9:\:\:\:\:\:\:\:\:\: update background model (see subection \ref{ssec:online adaptation mechanism}) \\
  \hline  
\end{tabular}

\subsection{Interesting Cases}
\subsubsection{Branches sway in the wind}
When, at some part of the scene, there are tree branches sway in the wind, the intensities of the pixels that depict this part will be clustered into two (or more) different clusters; intensities of the branches and intensities of the sky. In such cases, conventional methods that utilize a fixed number of components for representing the underlying data distribution may encounter serious problems. On the contrary, the proposed approach can estimate the number of components directly from the data. Therefore, the aforementioned clusters will be correctly identified and the corresponding pixels will be considered as background.
\subsubsection{Switching from background to foreground and vice versa}
Consider the following scenario; a foreground object, let us say a pedestrian, appears at pixel $x_i$ at time $t=t_0$ and stays there, standing still, before leaving at time $t=t_0+t_1$. Then, he/she returns back at the same location at time $t=t_0+t_1+t_2$. We want to discuss what will be the behavior of the proposed method regarding times $t_1$ and $t_2$. In order to provide a formal explanation, we have to employ i) the history of pixel's intensities, ii) relations Eq.(25), Eq.(27), Eq.(28) and Eq.(33), iii) the $p(bg)$ parameter and iv) the threshold for considering a pixel to belong to the background. We consider that the length of pixel’s  history equals $100$, the $\epsilon^*$ equals $2$ and the threshold for considering a pixel to belong to the background equals $0.5$. 

Consider that a pedestrian appears at pixel $x_i$ at frame $t_0$ (assuming that we use a camera with constant fps rate, we measure time in frames and not in seconds, this way the analysis is camera independent). A new component will be created for that pixel. The new component will be initialized using Eq.(28). Then, the pedestrian stays there (standing still) for the next frames. The following figures depict the evolution of the output of Eq.(33) using different values for $p(bg)$.

\begin{figure}[t]
	\begin{minipage}[b]{0.95\linewidth}
		\centering
		\centerline{\includegraphics[width=0.795\linewidth]{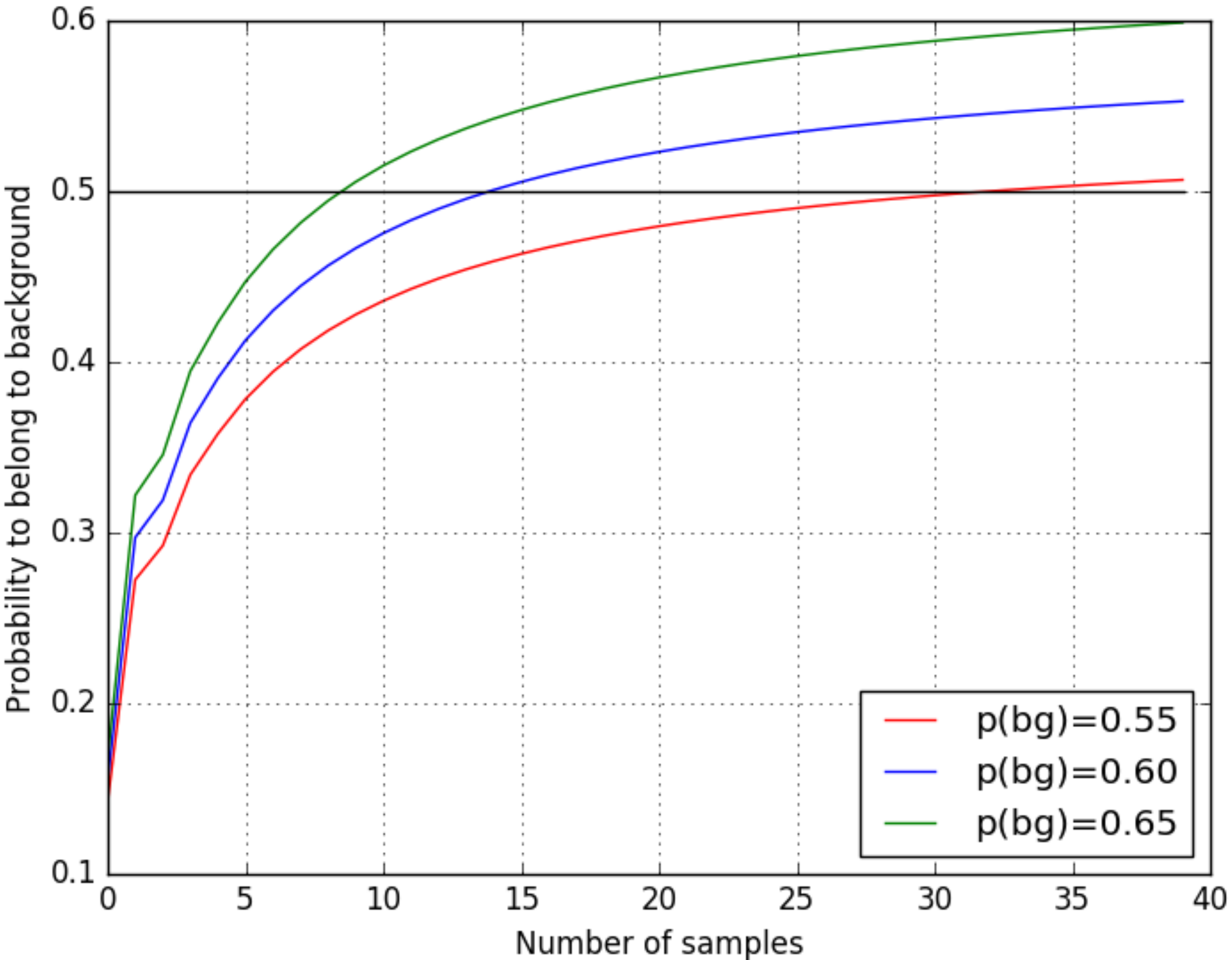}}
	\end{minipage}
	\caption{Time required for switching from foreground to background.} 
	\label{fig:switch}
\end{figure}

The $x-$axis in Fig.\ref{fig:switch} corresponds to the new captured frames. As we can see, if we use a value close to $0.5$ for $p(bg)$, then the pixel $x_i$ will be considered as background after 32 frames. If we increase the value for $p(bg)$ then much fewer frames are needed for considering the same pixel as background, because the prior belief that a pixel belongs to the background class is larger. 

Now consider that the pedestrian decides to leave this location. Then there are two different cases. In the first case, the pedestrian starts moving before he/she will be considered as background. In that case, the system will be correctly considering the pedestrian as foreground object for the whole period started at $t_0$. In the second case, the pedestrian decides to leave $N$ frames later after the time when pixel $x_i$ considered background. Since the rate of increment of the component's weight is the same as the rate of decrement, and due to the fact that $p(bg|x) + p(fg|x) = 1$, $N$ additional frames will be required before our system consider the pixel $x_i$ as background.

\subsubsection{Sensor noise and flickering intensities}
The sensor noise, typically, is zero mean, Gaussian and additive. In this case the noise will slightly affect the intensity of the pixels around the mean value. Therefore, variations due to this kind of noise will be captured by the components of the proposed GMM. On the other hand, flickering of intensities and/or salt and pepper sensor noise will indeed result to consider individual pixels as foreground. In this case we remove foreground blobs whose area is smaller than a threshold. During the evaluation of the proposed method, this threshold was manually optimized for each dataset and this post processing step applied on all algorithms that our method is compared against.

\section{In-Camera Acceleration Architecture}

In this section, we describe in detail the hardware architecture for the proposed background subtraction algorithm. We call the proposed parallel implementation as Background Subtraction Parallel System (BSPS). BSPS  exploits the reconfigurable resources of today's FPGA devices. 

\subsection{High Level Architecture}
Since the proposed approach makes no assumption for pixel relationships, the background model for all pixels can be computed independently. Thus, our hardware architecture consists of many parallel cores in a scalable configuration, each processing a single pixel.  In Section \ref{sec:experimental validation} we demonstrate two configurations; one low cost, featuring a 4-core BSPS Engine and a second one featuring a 16-core BSPS Engine.

\begin{figure}[t]
\begin{minipage}[b]{0.9\linewidth}
  \centering
  \centerline{\includegraphics[width=0.85\linewidth]{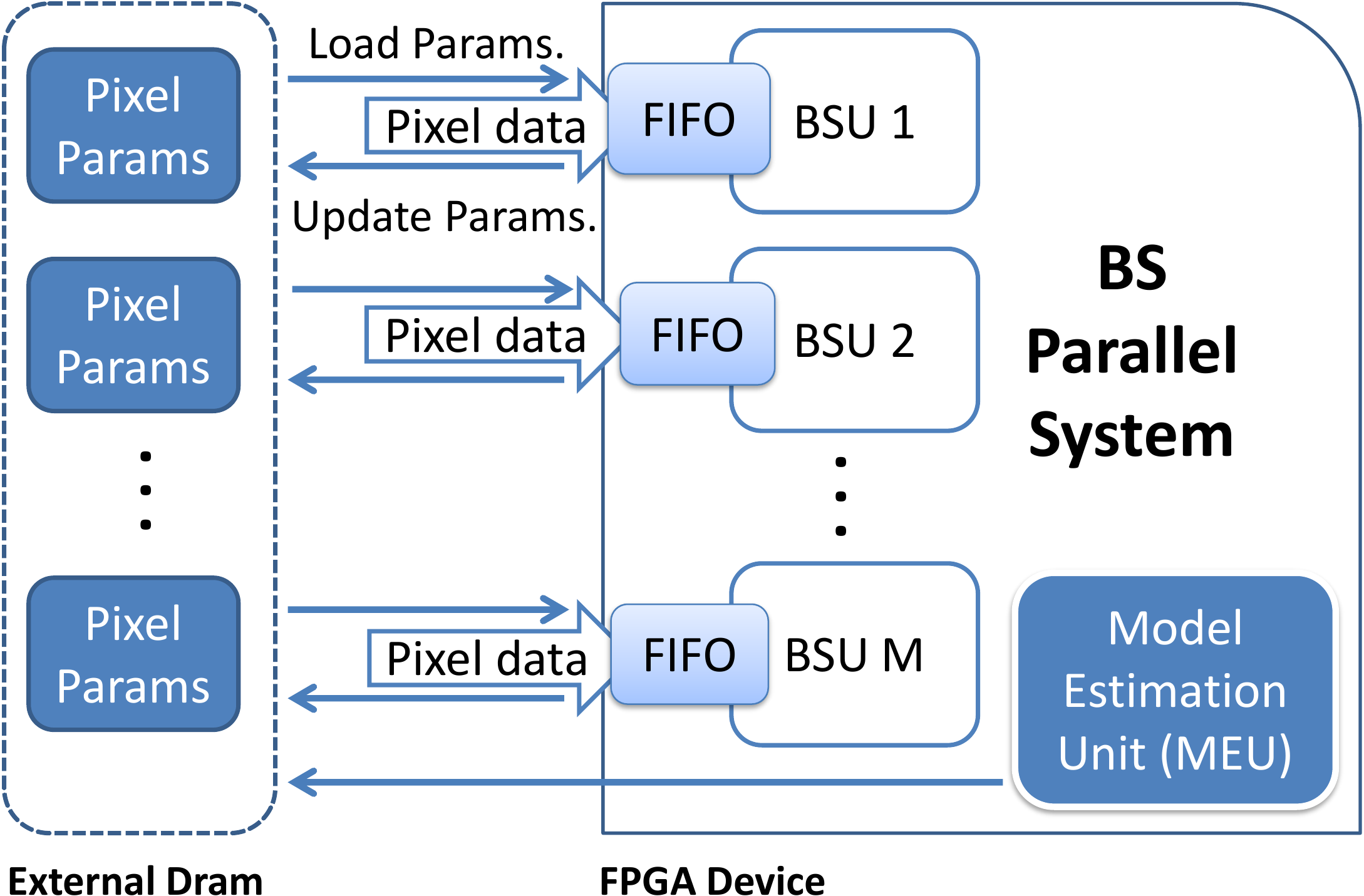}}
\end{minipage}
\caption{The data loading process of the BSPS.} 
\label{fig:BSPS}
\end{figure}

Each of the cores is connected to a shared bus in order to get the processing data from the external DRAM (or memory mapped camera module) of a host system. The data loading is performed in batches of up to 16 pixels as shown in Fig.\ref{fig:BSPS}.

All operations are per pixel with no dependencies between them. Thus, using a buffering scheme utilizing simple FIFOs, we can hide the latency of the external DRAM and make our scheme working seamlessly as a streaming accelerator. However, it has to be mentioned that the parallelization of the algorithm, or the overall performance in general, does not actually depend on the buffering scheme, which in our case prevents possible "data starvation" from the outside. The operations regarding data loading and write-back are fully pipelined. More details regarding the bandwidth demands are given in Section \ref{sec:experimental validation}. The output of each core is a probabilistic classification for the corresponding pixel (background or foreground) and the updated parameters of the background model.

\subsection{System Organization}
\begin{figure}[t]
	\begin{minipage}[b]{0.95\linewidth}
		\centering
		\centerline{\includegraphics[width=0.8\linewidth]{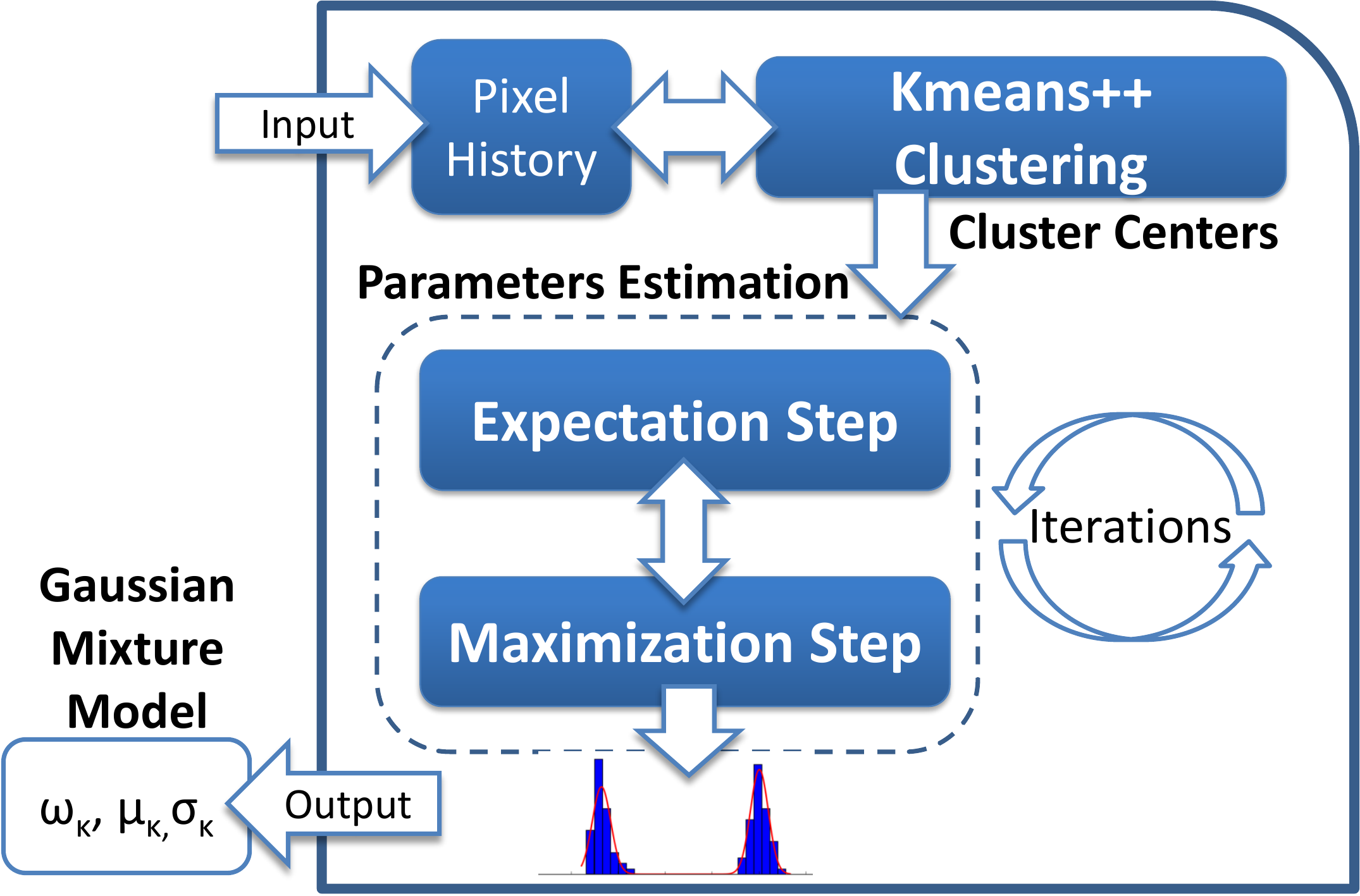}}
	\end{minipage}
	\caption{The Model Estimation Unit (MEU) organization.} 
	\label{fig:MEU}
\end{figure}

\begin{figure}[t]
	\begin{minipage}[b]{1.0\linewidth}
		\centering
		\centerline{\includegraphics[width=0.95\linewidth]{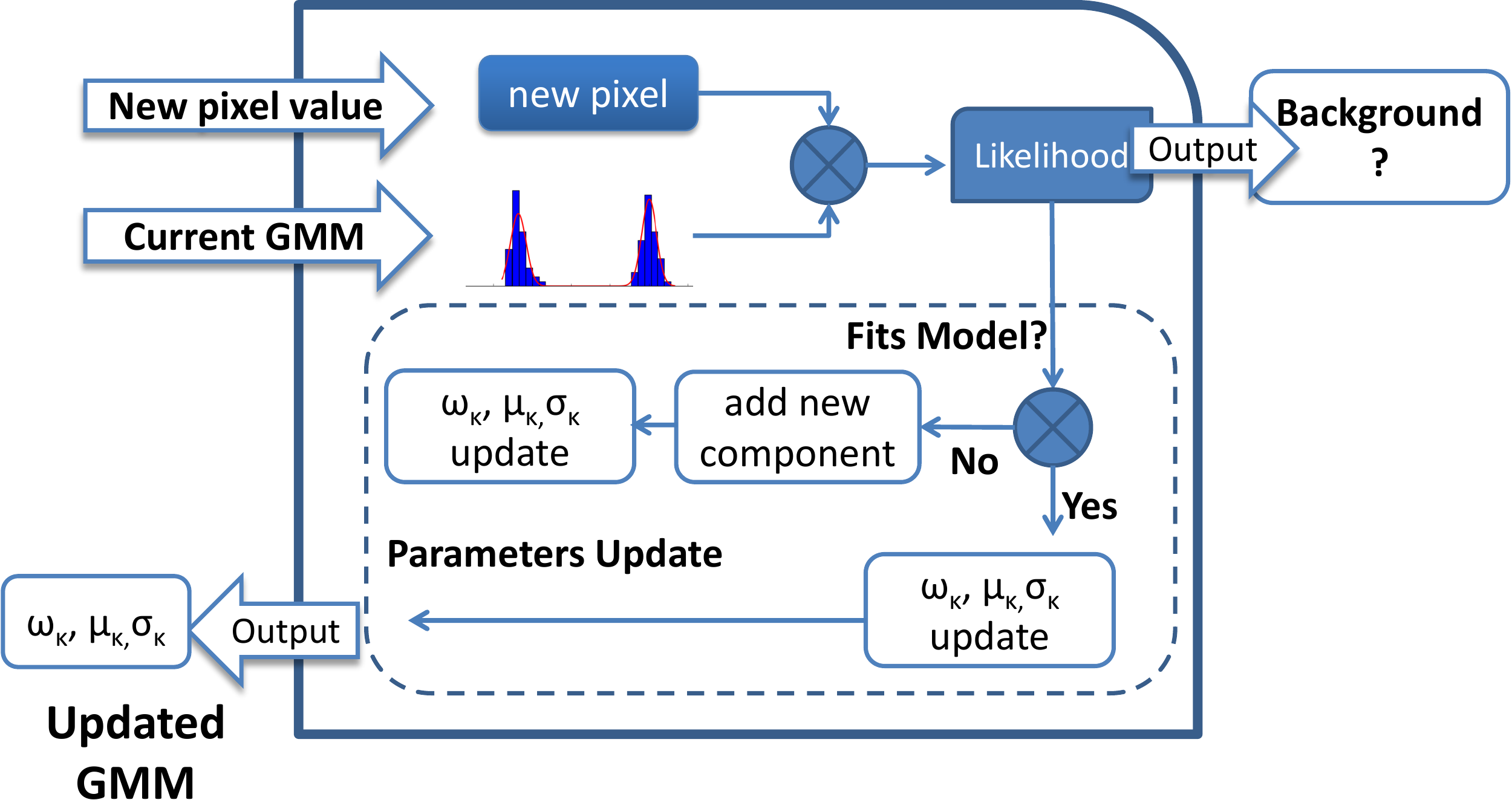}}
	\end{minipage}
	\caption{The Background Subtraction Unit (BSU) organization.} 
	\label{fig:BSU}
\end{figure}

The BSPS comprises of two basic sub-modules: the \textit{Model Estimation Unit} (MEU), which is depicted in Fig.\ref{fig:MEU} and the \textit{Background Subtraction Unit} (BSU) depicted in Fig.\ref{fig:BSU}. The MEU is activated just once at the initialization process of the system. It is responsible for building the proposed background model at all pixel locations. It uses a small history of pixel values and automatically estimates the appropriate number of Gaussian components along with their mixing coefficients. After the model is built, the MEU stores the model parameters to the external DRAM  for each one of the image pixels.

\label{sec:in-camera acceleration architecture}

Then and during the normal operation of the system, only the BSU is activated. The BSU takes as input the pixel data stream along with the model parameters and gives as output the probabilistic segmentation of each pixel, while it also updates model parameters. This way for each pixel a background model is maintained and updated, which is utilized for the classification of all the new incoming pixels.

\subsection{The Model Estimation Unit}   
One of the key advantages of the proposed scheme is that it does not require any prior knowledge about the structure of the background model in order to achieve an optimal operation. The MEU, depicted in Fig.\ref{fig:MEU}, is responsible for this task. It builds an accurate model for the specific background and its inherent temporal characteristics. It takes as input a small history of pixel responses ($\sim100$) at a specific pixel location and outputs an accurate background model for this pixel. As mentioned in Section \ref{sec:random variables optimization}, in this module two basic algorithms are utilized; the k-means++ and the EM algorithm. Around 100 frames correspond to 13 seconds of videos when the frame rate of the camera is 7.5Hz, a typical value for thermal cameras. It has to be mentioned that the presented algorithm does not require any reference background frame to operate properly. In case that foreground objects appear in these $\sim100$ frames, due to the fact that they are moving objects, they will be modeled by components with very low weight and thus they will slightly affect the background estimation process. Furthermore, by employing the updating mechanism the model will be adapted to new frames and will discard the components that model foreground objects. The history of frames could have been chosen to include 150 or 200 or even more frames. However, increasing the length of the history increases the computational requirements for model initialization. Since this work presents a model for in-camera background subtraction, $\sim100$ frames were chosen due to the fact that this number of frames is sufficient for describing the current dynamics of a scene and at the same time the computational cost for initializing the model is acceptable.  

\subsection{The Background Subtraction Unit}
The BSU, depicted in Fig.\ref{fig:BSU}, is responsible for classifying the incoming pixels into the two available classes (background and foreground) and also updating the background model according to the new pixel response. For this reason, BSU takes as input a new pixel value ($x_{new}$) and the current Gaussian mixture for this pixel, which is stored outside the chip, and gives as output the updated Gaussian mixture, as well as, the probabilistic classification of the incoming pixel. 

During the background subtraction task the FIFO based scheme processes all pixels of one frame before proceed to next frame; from the same frame it loads parallel batches of pixels depending on the number of parallel cores on chip. This way, we can achieve lower latency during processing and also have lower buffering when accessing the camera sensor. On the contrary, during the initialization of the system, the FIFO based scheme processes for each pixel a history of intensities, since it is required for the parameter estimation task.

\section{Experimental Validation}
\label{sec:experimental validation}

\subsection{VI Mixture Model Fitting Capabilities}  
During experimental validation, we evaluate the Variational Inference Mixture Model (VIMM) in terms of fitting accuracy and computational cost. We compare VIMM with the GMM and DPMM. The GMMs are employed under two different settings; (i) with more and (ii) less components than the underlying distribution. 
For experimentation purposes, we create one synthetic dataset 
from three non overlapping Gaussian distributions. 
The initial value for the number of components for VIMM is set to $10$. In order to compare our method with the conventional GMMs of fixed number of components, we create two Gaussian models of $10$ and $2$ components respectively. These numbers are arbitrarily chosen, since the correct number of components is not a priori known.  

\begin{figure}[t]
  \begin{minipage}{1.0\linewidth}
    \centering
    \centerline{\includegraphics[width=1.0\linewidth]{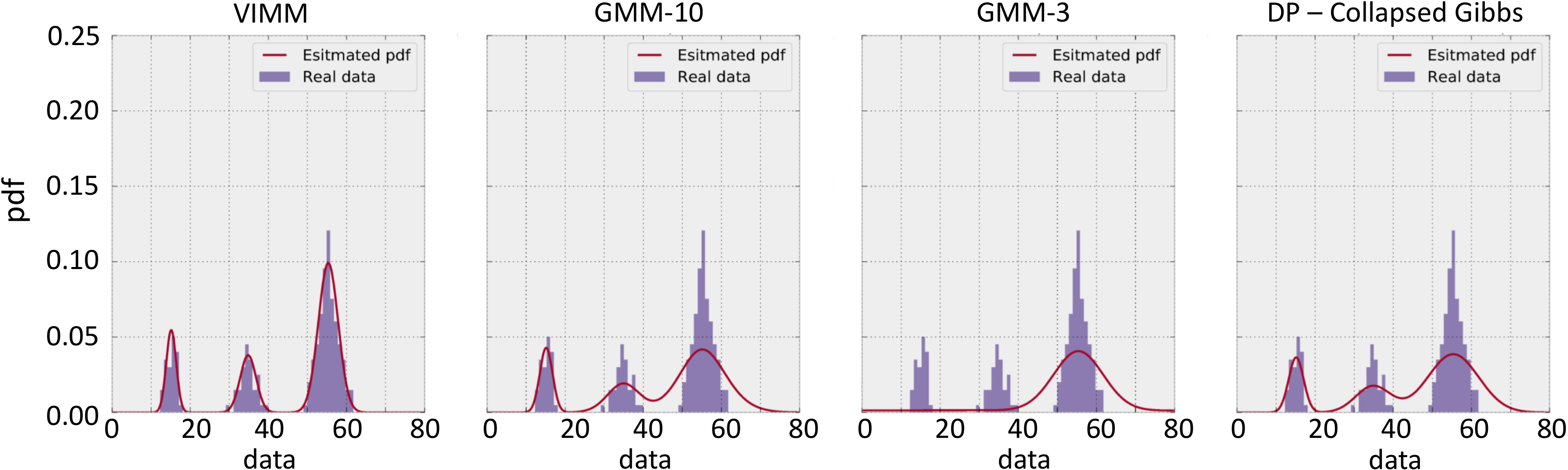}}
  \end{minipage}
  \caption{Fitting performance -- three Gaussian distributions.} 
    \label{fig:fig2}
\end{figure}

Fig. \ref{fig:fig2} presents the fitting performance of all models. Our method correctly estimates the number of components. The GMM with $2$ components under-fits the data, since the underlying distribution comes from $3$ Gaussians. The GMM with $10$ components also under-fits the data. This is very likely to happen when the underlying distribution of the data is simpler than the structure of the GMM. In such cases, the GMM is likely either to overfit the data by assigning some components to outliers or underfit the data by constructing several components to describe samples coming from the same Gaussian. The DPMM approach yields better results, since it is able to be adapted to the current data statistics, but it still under-fits the data. Table \ref{tab:tab1} presents time performance of the different models. All presented times were computed in Python and not in hardware implementation [see subsection \ref{ssec:hardware cost}].

\begin{table}[t]
\centering
\caption{Time performance of the different models in seconds.}
\label{tab:tab1}
\newcolumntype{L}[1]{>{\hsize=#1\hsize\raggedright\arraybackslash}X}%
\newcolumntype{C}[1]{>{\hsize=#1\hsize\centering\arraybackslash}X}%
\begin{tabularx}{1.0\linewidth}{L{7.5}C{4.0}C{4.0}C{4.0}C{4.0}}
\hline \hline
         & \vspace{0.005mm} \textbf{VIMM} & \vspace{0.005mm} \textbf{GMM-10} & \vspace{0.005mm} \textbf{GMM-2} & \vspace{0.005mm} \textbf{DPMM} \\
First dataset & 0.156  & 0.034 & 0.011 & 21.35  \\
Second dataset  & 0.124  & 0.067 & 0.031 & 30.19 \vspace{0.3cm}  \\ \hline \hline
\end{tabularx}
\end{table}

\subsection{Updating Mechanism Performance}
\label{ssec:Updating Mechanism Performance}
In this section we evaluate the quality of the proposed updating mechanism, with and without keeping in memory the observed data, and compare it against the updating mechanism presented in \cite{zivkovic_improved_2004}. The rationale behind the decision to explore both cases lies in the fact that we target special purpose hardware with very limited on-chip memory. In this respect, we have to validate that even without keeping the data in memory the algorithm performance is not affected at all.

\begin{figure}[t]
  \begin{minipage}{1.0\linewidth}
    \centering
        {\includegraphics[width=1.0\linewidth]{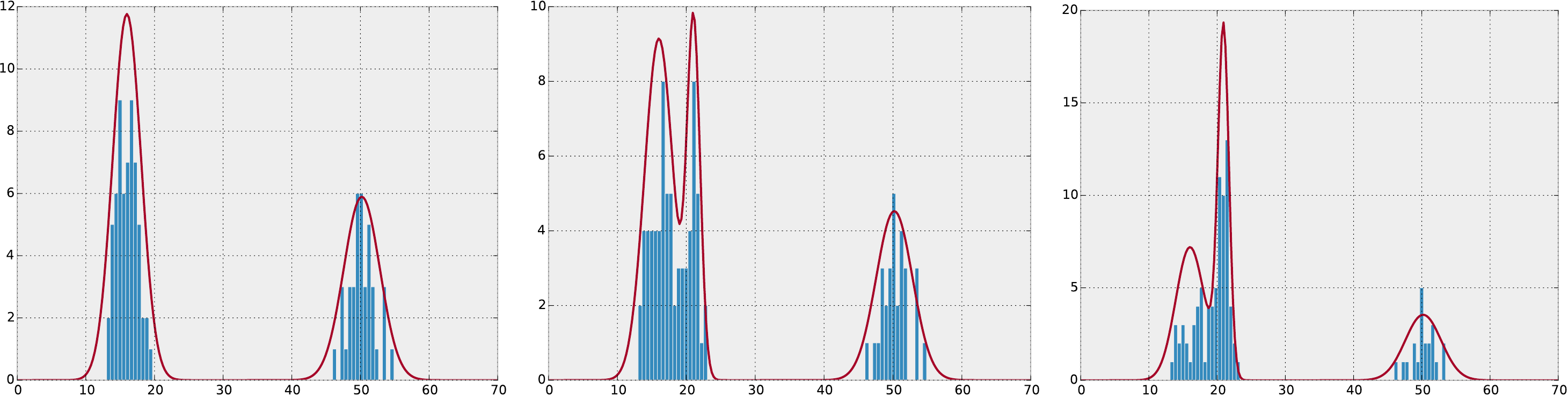}}
        \centerline{\footnotesize (a) Proposed adaptation process using observed data.} \vspace{0.001in}
  \end{minipage}
  \begin{minipage}{1.0\linewidth}
    \centering
            {\includegraphics[width=1.0\linewidth]{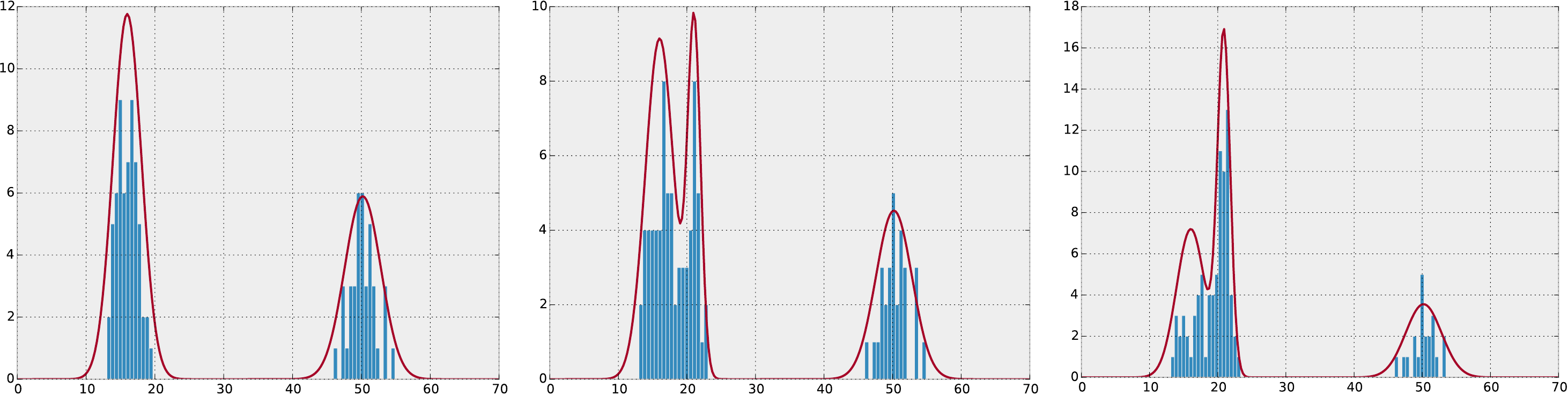}}
            \centerline{\footnotesize (b) Proposed adaptation process without using observed data.} \vspace{0.001in}
  \end{minipage}
  \begin{minipage}{1.0\linewidth}
      \centering
     {\includegraphics[width=1.0\linewidth]{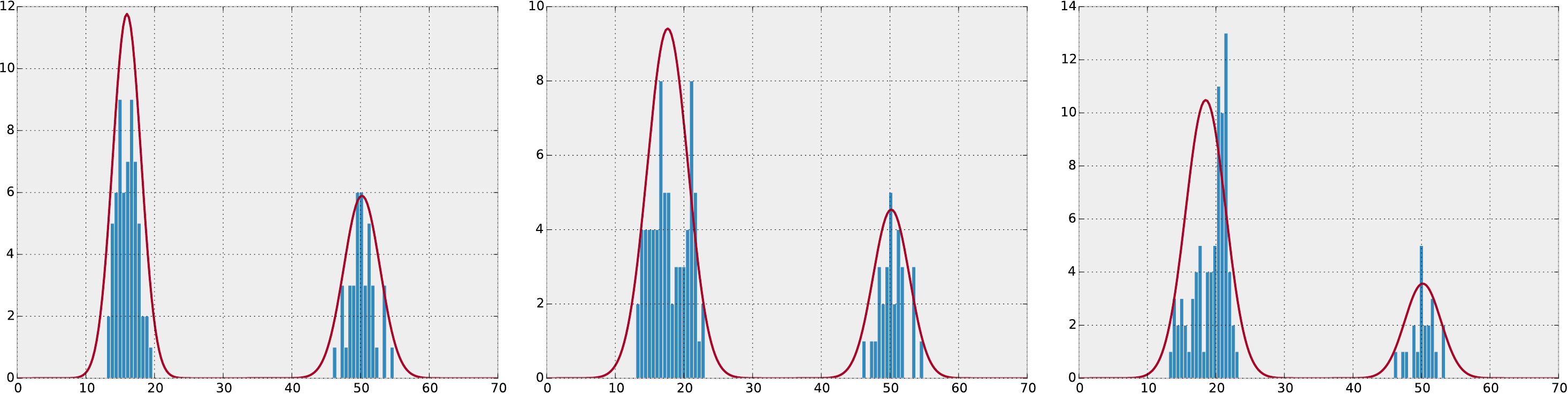}}
      \centerline{\footnotesize (c) Adaptation process presented in \cite{zivkovic_improved_2004}.}
    \end{minipage}
    \caption{Performance evaluation of model updating mechanisms.} 
     \label{fig:adaptation}
\end{figure}

Fig.\ref{fig:adaptation} presents the adaptation of all models. To evaluate the quality of the adaptation, we use a toy dataset with 100 observations. Observed data were generated from two Gaussian distributions with mean values 16 and 50 and standard deviations 1.5 and 2.0 respectively. The initially trained models are presented in the left column. Actually, there are two cases for evaluating the performance of the updating mechanisms. In the first case, the evaluation could have been performed by creating one more well-separated Gaussian. In the second one, which we have chosen to follow and is much harder, the performance is evaluated on a Gaussian distribution that overlaps with one of the two initial Gaussians. Therefore, we generated 25 new samples from a third Gaussian distribution with mean value 21 and standard deviation 1.0. The middle column indicates the adaptation performance after 25 new observations, while the right column after 50 new observations. Our model, either it uses the history of observed data or not, creates a new component and successfully fits the data. On the contrary, the model of \cite{zivkovic_improved_2004} is not able to capture the statistical relations of the new observations and fails to separate the data generated from the overlapping Gaussians (middle and right columns). The quality of the presented updating mechanism becomes more clear in the right column, which presents the adaptation of the models after 50 new observations. 

\subsection{Background Subtraction Algorithm Evaluation}
\subsubsection{OSU and AIA datasets}
For evaluating our algorithm, we use the Ohio State University (OSU) thermal datasets and a dataset captured at Athens International Airport (AIA) during eVacutate\footnote{http://www.evacuate.eu} European Union funded project. Specifically, we used two OSU datasets, referred as OSU1 and OSU2, which contain frames that have been captured using a thermal camera and have been converted to grayscale images. On the contrary, the AIA dataset contains raw thermal frames whose pixel values correspond to the real temperature of objects.  

OSU datasets \cite{davis_fusion-based_2005, davis_robust_2004, davis_background-subtraction_2007} are widely used for benchmarking algorithms for pedestrian detection and tracking in thermal imagery. Videos were captured under different illumination and weather conditions. AIA dataset was captured using a Flir A315 camera at different Airside Corridors and the Departure Level. Ten video sequences were captured, with frame size $320 \times 240$ pixels of total duration 32051 frames, at 7.5fps. The experimentation was conducted throughout the third pilot scenario of eVacuate project. For all datasets we created our own ground truth by selecting 50 frames randomly but uniformly distributed, in order to cover the whole videos duration. Then, we manually annotated this frames by creating a binary mask around the foreground objects.

We compared our method against the method of \cite{zivkovic_improved_2004} (MOG), which is one of the most robust and widely used background subtraction techniques. MOG algorithm uses a pre-defined number of Gaussian components for building the background model. In order to perform a fair comparison we fine-tune the parameters of MOG algorithm for each of the two datasets to optimize its performance. Furthermore, we compare our method against the method of  \cite{davis_robust_2004, davis_background-subtraction_2007} (SBG) 
used for background substraction in thermal data. This method uses a single Gaussian distribution for modeling the background and, thus, it often under-fits the data. Comparison against this technique can highlight problems that arise when the number of components of a GMM is underestimated. We do not compare our method against a DPMM-based background subtraction technique, like the one in \cite{haines_background_2014}, since its computational cost is high and we target low power and memory devices.

\begin{figure}
\begin{minipage}[b]{1.0\linewidth}
  \centering
  \centerline{\includegraphics[width=1.0\linewidth]{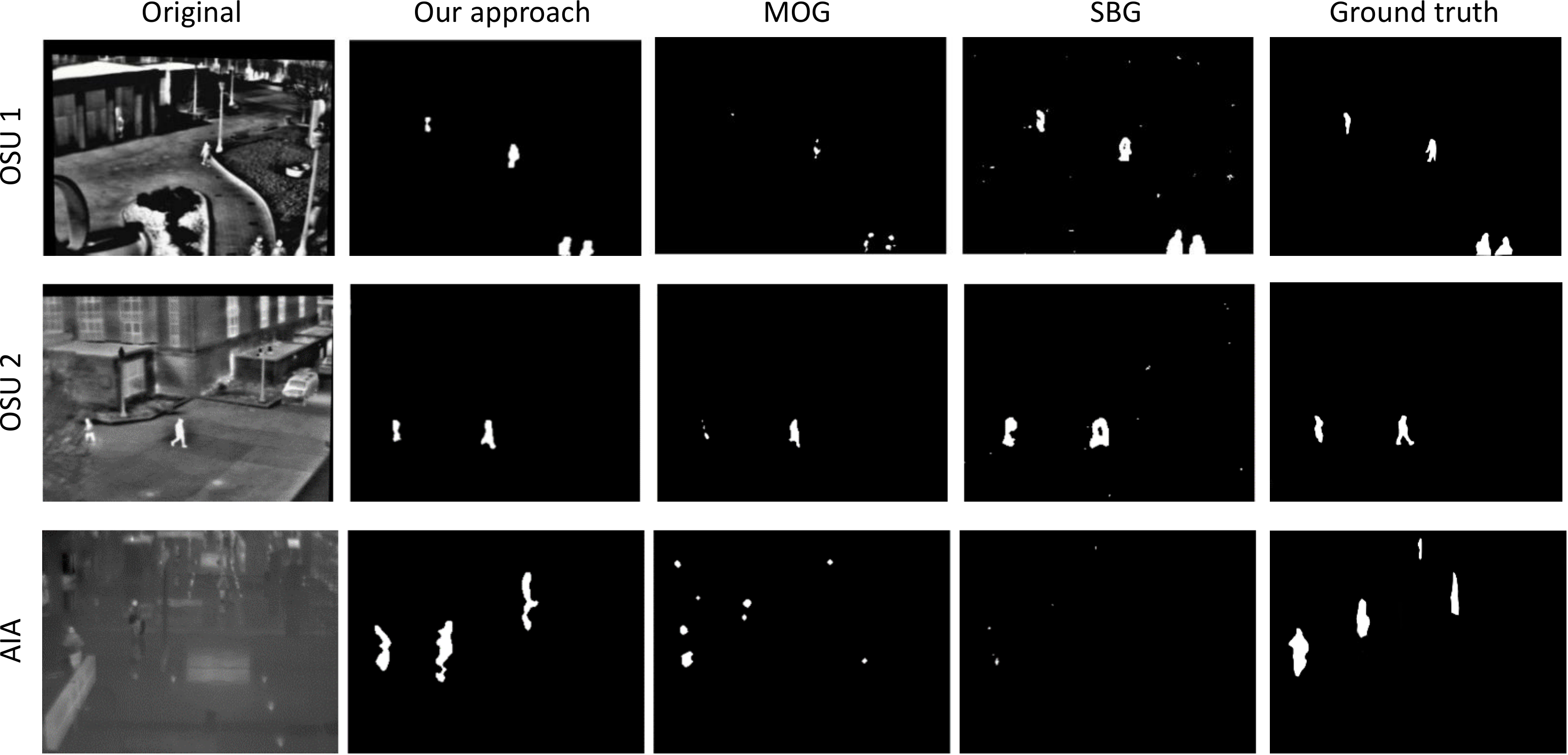}}
\end{minipage}
\caption{Visual results for all datasets.}
\label{fig:OSU4}
\end{figure}

\begin{figure}[t]
\begin{minipage}[b]{1.0\linewidth}
  \centering
  \centerline{\includegraphics[width=0.95\linewidth]{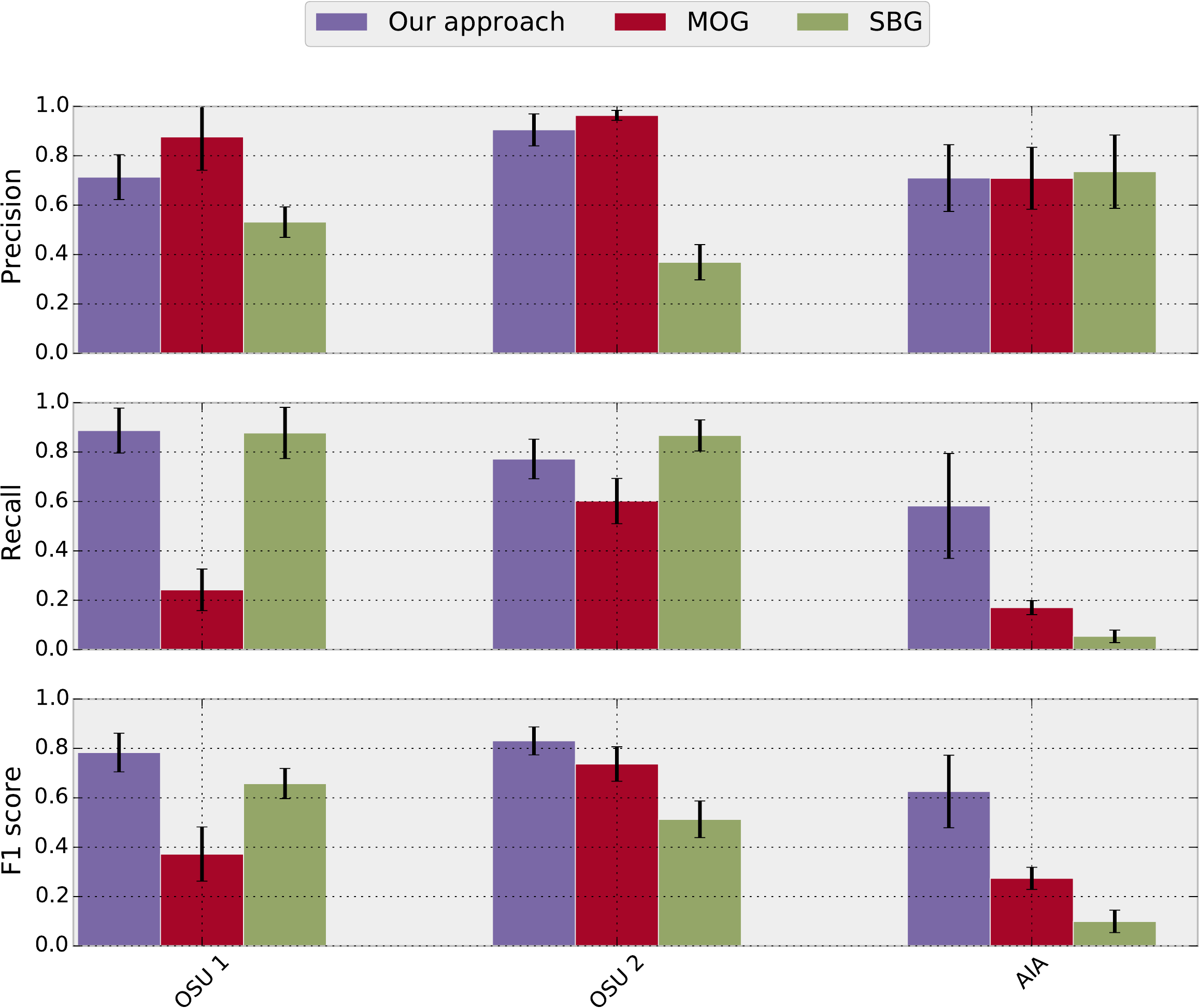}}
  \centerline{\footnotesize Precision, recall and F1 score} \medskip
\end{minipage}
\caption{Algorithms performance per dataset.}
\label{fig:prf}
\end{figure}

For estimating the background model, we utilized 100 frames as history and set the maximum number of components equal to 50. After the initialization of the background model, we observed that the models for OSU and AIA datasets consists of $2$ to $4$ and $3$ to $6$ components respectively. Fig. \ref{fig:OSU4} visually present the performance of the three methods. As is observed, our method outperforms both MOG and SBG on all datasets. While MOG and SBG perform satisfactory on grayscale frames of OSU datasets, their performance collapses when they applied on AIA dataset, which contains actual thermal responses, due to their strong assumptions regarding the distribution of the responses of pixels and the peculiarities of thermal imagery i.e. high signal-to-noise ratio, lack of color and texture and non-homogeneous thermal responses of objects (see Section I). 
Then, an objective evaluation takes place in terms of \textit{recall}, \textit{precision} and \textit{F1 score}. Regarding OSU datasets, MOG algorithm presents high precision, however, it yields very low recall values, i.e. the pixels that have been classified as foreground are indeed belong to the foreground class, but a lot of pixels that in fact belong to background have been misclassified. SBG algorithm seems to suffer by the opposite problem. Regarding AIA dataset, our method significantly outperforms both approaches. Although, MOG and SBG algorithms present relative high precision, their recall values are under $20\%$. Figure \ref{fig:prf} presents average precision, recall and F1 score per dataset and per algorithm.

Regarding computational cost, the main load of our algorithm is in the implementation of EM optimization. In all experiments conducted, the EM optimization converges within 10 iterations. Practically, the time required to apply our method is similar to the time requirements of Zivkovic's method making it suitable for real-time applications.

\subsubsection{Change Detection Benchmark}
Besides OSU and AIA datasets we evaluated the performance of our algorithm on the \textit{change detection benchmark - 2014} (CDB). The CDB provides five thermal videos, recorded at indoor and outdoor environments,  along with their ground truth. For evaluating the performance of our algorithm, we utilized the same metrics as in CDB-2014, i.e. \textit{precision}, \textit{recall} and \textit{F1 score}, \textit{specificity}, \textit{False Positive Rate} (FPR), \textit{False Negative Rate} (FNR) and \textit{Percentage of Wrong Classifications} (PWC).

\begin{table}[t]
\centering
\caption{Performance evaluation on the thermal datasets of CDB.}
\label{tab:perf}
\newcolumntype{L}[1]{>{\hsize=#1\hsize\raggedright\arraybackslash}X}%
\newcolumntype{C}[1]{>{\hsize=#1\hsize\centering\arraybackslash}X}%
\begin{tabularx}{1.0\linewidth}{L{7.73}C{1.63}C{1.63}C{1.63}C{1.63}C{1.63}C{1.63}C{1.63}}
\hline \hline
\vspace{0.001mm} \textbf{Method}        & \vspace{0.001mm} \textbf{Prec.} & \vspace{0.001mm} \textbf{Rec.} & \vspace{0.001mm} \textbf{F1S} & \vspace{0.001mm} \textbf{Spec.} & \vspace{0.001mm} \textbf{FPR} & \vspace{0.001mm} \textbf{FNR} & \vspace{0.001mm} \textbf{PWC} \\
Our Method                               & 0.718  & 0.868  & 0.732 & 0.970 & 0.031 & 0.132 & 3.347 \\
Cascade CNN \cite{wang2016interactive}   & \textbf{0.951}  & \textbf{0.899}  & \textbf{0.920} & \textbf{0.997} & \textbf{0.003} & \textbf{0.049} & \textbf{0.405} \\
DeepBS \cite{babaee2017deep}             & 0.754  & 0.833  & 0.746 & 0.990 & 0.009 & 0.245 & 1.992 \\
IUTIS-5 \cite{bianco2015far}             & 0.785  & 0.808  & 0.771 & 0.994 & 0.005 & 0.215 & 1.198 \\ 
SuBSENSE \cite{st2015subsense}           & 0.812  & 0.751  & 0.747 & 0.990 & 0.009 & 0.187 & 1.678 \\
PAWCS \cite{st2015self}                  & 0.772  & 0.785  & 0.750 & 0.995 & 0.005 & 0.228 & 1.199 \\ 
\hline \hline
\end{tabularx}
\end{table}

Table \ref{tab:perf} presents the performance of our algorithm on the thermal datasets of CDB and compares it against the top two methods for each one of the metrics. The method of \cite{wang2016interactive} outperforms all methods. Our method presents the second highest recall, however, due to its lower precision it presents lower F1 score. Although, our method performs slightly worst than the leaders of CDB 2014, it is much less complicated and thus suitable for implementation in-camera.

\subsection{Hardware Cost}
\label{ssec:hardware cost}

The main argument of this work is that a novel highly accurate and demanding algorithm that needs to run in a 24/7 basis could be handled very efficiently by a reconfigurable device running as an in-camera accelerator. Thus, we primarily demonstrate our system in a low cost Xilinx Atrix7 FPGA device (xc7a200tfbg484-3). In addition, we deploy our system in a more powerful Virtex7 device (xc7vx550tffg1158-3) to show that it seamlessly scales to support more parallel cores.
For the code synthesis and bitstream generation we used Xilinx Vivado and Vivado HLS. For validation and proof only purposes our system was implemented in a low end Zedboard evaluation platform 
powered by a small Xilinx Zynq device.

Table \ref{tab:tab2} shows the hardware utilization for the Artix7 device when implementing 4 BSU cores and 1 MEU core. We implemented only 1 MEU, since this unit operates only for the initialization and parameter estimation of the system, and thus, its performance is not crucial. Table \ref{tab:tab2} also shows that the critical resource is the LUTs and DSPs. This is justified by the fact that the operations involved in the algorithm are mostly multiplications and divisions, and thus, apart from the DSPs, additional logic and signals are necessary to route the intermediate results and handle all algorithm states. DRAM utilization is almost zero as all operations are per pixel and no further caching in data is necessary, since there is no need for keeping the observed data in memory. It should be mentioned that keeping the observed data in memory would prohibit the implementation of this algorithm in low memory devices.
Table \ref{tab:tab3} shows the hardware utilization of the Virtex 7 device when implementing 16 BSU cores and 1 MEU core. The resource utilization in this case follows the same reasoning as before. The above two hardware configurations are compared with a quad-core ARM Cortex A9 CPU (Exynos4412 SoC) clocked at 1.7 GHz with 2GB RAM and a low power mobile Intel i5 (2450M) Processor clocked at 2.5Ghz with 8GB RAM, which features two physical cores with hyper threading capability (4 threads in total). It is selected for the evaluation as it offers a competitive computation power per watt.

\begin{table}[t]
\centering
\caption{Typical Hardware Cost on low cost, low power Xilinx Artix7 Device (xc7a200tfbg484-3). 4-BSU cores/1-MEU core.}
\label{tab:tab2}
\newcolumntype{L}[1]{>{\hsize=#1\hsize\raggedright\arraybackslash}X}%
\newcolumntype{C}[1]{>{\hsize=#1\hsize\centering\arraybackslash}X}%
\begin{tabularx}{1.0\linewidth}{L{13.0}C{4.0}C{4.0}C{4.0}}
\hline \hline
\vspace{0.005mm} \textbf{Logic Utilization}         & \vspace{0.005mm} \textbf{Used} & \vspace{0.005mm} \textbf{Available} & \vspace{0.005mm} \textbf{Utilization} \\
Number of Flip Flops  & 143089  & 269200 & 53\% \\
Number of Slice LUTs  & 119964  & 129000 & 92\%   \\
Number of DSP48E & 506  & 740 & 68\%  \\
Number of Block RAM\_18K  & 20  & 730 & 2\%    \\ \hline \hline
\end{tabularx}
\end{table}

\begin{table}[t]
\centering
\caption{Typical Hardware Cost on Xilinx Virtex 7 device (xc7vx550tffg1158-3). 16-BSU cores/1-MEU core.}
\label{tab:tab3}
\newcolumntype{L}[1]{>{\hsize=#1\hsize\raggedright\arraybackslash}X}%
\newcolumntype{C}[1]{>{\hsize=#1\hsize\centering\arraybackslash}X}%
\begin{tabularx}{1.0\linewidth}{L{13.0}C{4.0}C{4.0}C{4.0}}
\hline \hline
\vspace{0.005mm} \textbf{Logic Utilization}         & \vspace{0.005mm} \textbf{Used} & \vspace{0.005mm} \textbf{Available} & \vspace{0.005mm} \textbf{Utilization} \\
Number of Flip Flops  & 241604  & 692800 & 35\% \\
Number of Slice LUTs  & 269004  & 346400 & 78\%   \\
Number of DSP48E & 1184  & 2880 & 41\%  \\
Number of Block RAM\_18K  & 14.50  & 1180 & 1.2\%    \\ \hline \hline
\end{tabularx}
\end{table}

\begin{table}[t]
\centering
\caption{Comparison table between a Xilinx Atrix7 device @210Mhz, a Xilinx Virtex7 device @ 222 Mhz, an Intel i5 @2.5Ghz an ARM Cortex A9 @1.7Ghz and a DSP @ 600Mhz.}
\label{tab:tab4}
\newcolumntype{L}[1]{>{\hsize=#1\hsize\raggedright\arraybackslash}X}%
\newcolumntype{C}[1]{>{\hsize=#1\hsize\centering\arraybackslash}X}%
\begin{tabularx}{1.0\linewidth}{C{12.5}C{4.7}C{4.7}C{3.1}}
\hline \hline
 \\ \textbf{Image frame} & $\bf{320\times240}$ & $\bm{640\times480}$ &$\bm{\mu}$\textbf{J/pixel}\\
  \textbf{Artix 7} 4--cores & 17.36 fps & 4.34 fps & 3.45 \\
  \textbf{Vertex 7} 16-cores & 69.88 fps & 17.47 fps & 3.49 \\
  \textbf{ARM A9} 4-cores & 8.27 fps & 2.07 fps & 4.7-6.2 \\ 
  \textbf{Intel i5} 2-cores/ 4-threads & 58.59 fps & 14.56 fps & 5.82 \\
  \textbf{MOG \cite{zivkovic_improved_2004}} BF-537 DSP & 3.57 fps & - & -
\\ \hline \hline
\end{tabularx}
\end{table}

For the Intel i5 the software compiler platform used was Microsoft Visual Studio 2012 and our code was optimized for maximum speed (-O2 optimization level). For the ARM A9 platform, we used a lightweight XUbuntu 13.10 operating system with a g++ compiler using -O2 and-O3 optimization level. 
In all software reference implementations OpenMP was also used to utilize all the available cores/threads of the platform.
For the FPGA, we measure the exact clock cycles needed for segmenting a single pixel by a single core including loading and write back cycles. For this purpose, we use the Zedboard evaluation board. The exact clock cycles measured between 700-830 when real datasets are evaluated. These measurements are also verified for the proposed Atrix7 and Virtex7 devices using post-place and route timing simulation.

The I/O latency between the DRAM and the FPGA is completely hidden as the operations for each core are depending only on a single pixel and its corresponding background model. All this information is encoded in about 256 bits in average, thus a buffering scheme using simple FIFOs is easily implemented. The bandwidth demands between the device and the DRAM is no more than 250 MB/sec for 25FPS at 640x480 resolution, which is easily achievable even from low-end FPGA devices.
In all the experiments for the Intel i5 and ARM A9 we start measuring latency times after the data are loaded in the DRAM of the CPU. This probably is in benefit of the CPUs as the cache is hot in most of the measured cases. 

Table \ref{tab:tab4} shows that implementing just 4-cores in the Atrix7 device we get 17.3 FPS at 320x240 exceeding by far the capabilities of the FLIR A-315 thermal camera. The 4-core FPGA design outperforms the ARM A9 quad core CPU giving twice the FPS. In terms of power, Atrix7 consumes 4.6 watts based on Vivado's Power analysis tool while quad core ARM A9 consumes about 3.5-4 watts 
As expected the Intel i5 utilizing 4-threads outperforms the two previous platforms offering also the best performance per core. Its consumption is measured at 26.2 watts 
and refers only to the CPU consumption. The Virtex 7 device offers better performance, as it is capable of fitting 16-BSU cores. In terms of power the Virtex7 consumes 18.6 Watts measured using Vivado's Power analysis tool. 

Looking at the energy metric μJ/pixel in Table \ref{tab:tab4}, both FPGA devices give similar μJ/pixel and also better than the Intel i5. For the ARM A9 this metric is expressed as a range as it is based mostly on specs. In our evaluation experiments we could measure the total dynamic power of the board using the ODROID smart Power5 but it is not possible to safely measure only the CPU core modules.

The last column in Table \ref{tab:tab4} refers to the work of \cite{shen2012efficient} which implements the original MOG algorithm in an in-camera DSP processor (Blackfin BF-537) as a reference design for his proposed scheme. Even though it is hard to make a direct comparison, we see how challenging for embedded processors is it to keep up with the demanding task of background segmentation; even for a less accurate algorithm such as MOG.


\section{Conclusions}
\label{sec: conclusions}
In this work a novel algorithm for background subtraction was presented which is suitable for in-camera acceleration in thermal imagery. The presented scheme through an automated parameter estimation process, takes into account the special characteristics of data, and gives highly accurate results without any fine-tuning from the user. It is implemented in reconfigurable hardware using a HLS design flow with no approximations in accuracy, arithmetic or in the mathematical formulation of the proposed algorithm. Unlike previously introduced custom-fit hardware accelerators, our scheme is platform independent, scalable and easily maintainable. Finally, to the best of our knowledge this is the first time that the very demanding task of background subtraction can be executed to thermal camera sensor in real-time and at low power budget, which allows for a distributed new approach that avoids the bottlenecks of the existing centralized solutions.


\appendices
\section{Derivation of Optimal Variational Distributions}
\label{ap:appendix}
Using (\ref{eq:q_star}) and (\ref{eq:joint_factorization}) the logarithm of $q^*(\bm Z)$ is given by
\begin{equation}
\begin{aligned}
\ln q^*(\bm Z) = & \mathbb{E}_{\bm \varpi}[\ln p(\bm Z|\bm \varpi)] + \\ & + \mathbb{E}_{\bm \mu, \bm \tau}[\ln p(\bm X|\bm Z, \bm \mu, \bm \tau)] + \mathcal{C}
\end{aligned}
\label{eq:q_Z_optimized_2}
\end{equation}
substituting (\ref{eq:p_Z}) and (\ref{eq:p_X}) into (\ref{eq:q_Z_optimized_2}) we get
\begin{subequations}
\begin{align}
& \ln q^*(\bm Z)=  \sum_{n=1}^{N} \sum_{k=1}^{K} z_{nk} \bigg( \mathbb{E}\big[\ln \varpi_k \big] + \frac{1}{2}\mathbb{E}\big[\ln \tau_k\big] -\nonumber \\ 
&\:\:\:\:\:\:\:\:\:\:-\frac{1}{2}\ln2\pi - \frac{1}{2} \mathbb{E}_{\bm \mu, \bm \tau}\big[(x_n-\mu_k)^2\tau_k \big]\bigg) + \mathcal{C} \Rightarrow \nonumber  
\end{align}
\label{eq:q_star_Z_derivation}
\end{subequations}  


Using (\ref{eq:joint_factorization}) and (\ref{eq:q_star}) the logarithm of $q^*(\bm \varpi, \bm \mu, \bm \tau)$ is
\begin{subequations}
\begin{align}
\ln q^*(\bm \varpi, \bm \mu, \bm \tau) & =  \mathbb{E}_{\bm Z}\big[\ln p(\bm X|\bm Z, \bm \mu, \bm \tau) + \nonumber \\ 
& + \ln p(\bm Z|\bm \varpi) + \nonumber \\ 
& + \ln p(\bm \varpi) + \ln p(\bm \mu, \bm \tau)\big] + \mathcal{C} = \\
& = \sum_{n=1}^{N}\sum_{k=1}^{K} \mathbb{E}\big[z_{nk}\big] \ln \mathcal{N}(x_n|\mu_k, \tau_k^{-1}) + \nonumber \\ 
&+ \mathbb{E}_{\bm Z}\big[\ln p(\bm Z|\bm \varpi)\big] \nonumber \\
& + \ln p(\bm \varpi) + \sum_{k=1}^{K}\ln p(\mu_k, \tau_k) + \mathcal{C}
\label{eq:q_varpi_mu_tau_optimized_b}
\end{align}
\label{eq:q_varpi_mu_tau_optimized}
\end{subequations} 
Due to the fact that there is no term in (\ref{eq:q_varpi_mu_tau_optimized_b}) that contains parameters from both sets $\{\bm \varpi\}$ and $\{\bm \mu, \bm \tau\}$, the distribution $q^*(\bm \varpi, \bm \mu, \bm \tau)$ can be factorized as $q(\bm \varpi, \bm \mu, \bm \tau) = q(\bm \varpi) \prod_{k=1}^{K}q(\mu_k, \tau_k)$.
The distribution for $q^*(\bm \varpi)$ is derived using only those terms of (\ref{eq:q_varpi_mu_tau_optimized_b}) that depend on the variable $\bm \varpi$. Therefore the logarithm of $q(\bm \varpi)$ is given by
\begin{subequations}
\begin{align}
\ln q^*(\bm \varpi) & = \mathbb{E}_{\bm Z}\big[\ln p(\bm Z|\bm \varpi)\big] + \ln p(\bm \varpi) + \mathcal{C} = \\
& = \sum_{k=1}^{K} \ln \varpi_k^{(\sum_{n=1}^{N}r_{nk} + \lambda_0 -1)} + \mathcal{C} = \\
& = \sum_{k=1}^{K} \ln \varpi_k^{(N_k + \lambda_0 -1)} + \mathcal{C}
\label{eq:q_star_varpi_derivation_c}
\end{align}
\end{subequations}  
We have made use of $\mathbb{E}[z_{nk}]=r_{nk}$, and we have denote as $N_k=\sum_{n=1}^{N}r_{nk}$. (\ref{eq:q_star_varpi_derivation_c}) suggests that $q^*(\bm \varpi)$ is a Dirichlet distribution with hyperparameters $\bm \lambda = \{N_k + \lambda_0\}_{k=1}^K$.

Using only those terms of (\ref{eq:q_varpi_mu_tau_optimized_b}) that depend on variables $\bm \mu$ and $\bm \tau$, the logarithm of $q^*(\mu_k, \tau_k)$ is given by

\begin{align}
\ln q^*(\mu_k, \tau_k) & = \ln \mathcal{N}(\mu_k|m_0, (\beta_0 \tau_k)^1) + \nonumber \\ 
&\:\:\:\:\: + \ln Gam(\tau_k|a_0, b_0) + \nonumber \\
&\:\:\:\:\: + \sum_{n=1}^{N}\mathbb{E}\big[z_{nk}\big]\ln \mathcal{N}(x_n|\mu_k,\tau_k^{-1}) + \mathcal{C} = \nonumber \\ 
& = -\frac{\beta_0\tau_k}{2}(\mu_k-m_0)^2 + \frac{1}{2}\ln (\beta_0 \tau_k) + \nonumber \\ 
&\:\:\:\:\: + (a_0-1)\ln \tau_k - b_0\tau_k - \nonumber \\ &\:\:\:\:\: -\frac{1}{2}\sum_{n=1}^{N}\mathbb{E}\big[z_{nk}\big](x_n-\mu_k)^2\tau_k + \nonumber \\ 
&\:\:\:\:\: + \frac{1}{2}\bigg(\sum_{n=1}^{N}\mathbb{E}\big[z_{nk}\big] \bigg) \ln(\beta_0\tau_k) + \mathcal{C}
\label{eq:q_star_mu_tau_derivation_b}
\end{align} 
For the estimation of $q^*(\mu_k|\tau_k)$, we use (\ref{eq:q_star_mu_tau_derivation_b}) and keep only those factors that depend on $\mu_k$.
\begin{subequations}
\begin{align}
\ln q^*(\mu_k|\tau_k) & = -\frac{\beta_0\tau_k}{2}\big(\mu_k-m_0\big)^2 - \nonumber \\
&\:\:\:\:\: - \frac{1}{2}\sum_{n=1}^{N}\mathbb{E}\big[z_{nk}\big]\big(x_n - \mu_k\big)^2\tau_k = \\
& = -\frac{1}{2}\mu_k^2\Big(\beta_0 + N_k\Big)\tau_k +\nonumber \\ 
&\:\:\:\:\: + \mu_k \tau_k \Big(\beta_0 m_0 + N_k\bar x_k\Big) + \mathcal{C} \Rightarrow \\
& q^*(\mu_k|\tau_k) = \mathcal{N}(\mu_k|m_k, (\beta_k \tau)^{-1})
\label{eq:q_star_mu_derivation_b}
\end{align}
\label{eq:q_star_mu_derivation}
\end{subequations}  
where $\bar x_k = \frac{1}{N_k}\sum_{n=1}^{N}r_{nk}x_n$, $\beta_k = \beta_0 + N_k$ and $m_k = \frac{1}{\beta_k}(\beta_0 m_0 + N_k \bar x_k)$. 

After the estimation of $q^*(\mu_k|\tau_k)$, logarithm of the optimized the distribution $q^*(\tau_k)$ is given by
\begin{subequations}
\begin{align}
\ln q^*(\tau_k) & = \ln q^*(\mu_k, \tau_k) - \ln q^*(\mu_k|\tau_k) = \\
& = \bigg(a_0+\frac{N_k}{2} - 1\bigg) \ln \tau_k - \nonumber \\ &\:\:\:\:\:\: -\frac{1}{2}\tau_k\bigg(\beta_0\big(\mu_k-m_0\big)^2 + \nonumber \\
&\:\:\:\:\:\: +2b_0 + \sum_{n=1}^{N}r_{nk}\big(x_n-\mu_k\big)^2 -\nonumber \\ &\:\:\:\:\:\: -\beta_k\big(\mu_k-m_k\big)^2 \bigg) + \mathcal{C} \Rightarrow \\
& q^*(\tau_k) = Gam(\tau_k|a_k, b_k)
\label{eq:q_star_tauk_derivation_b}
\end{align}
\label{eq:q_star_tauk_derivation}
\end{subequations}
The parameters $a_k$ and $b_k$ are given by 
\begin{subequations}
\begin{align}
a_k & = a_0 +  \frac{N_k}{2} \\
b_k & = b_0 + \frac{1}{2}\bigg(N_k\sigma_k + \frac{\beta_0 N_k}{\beta_0 + N_k}\big(\bar x_k - m_0\big)^2 \bigg)
\end{align}
\label{eq:ap_q_star_tauk}
\end{subequations}
where $\sigma_k = \frac{1}{N_k}\sum_{n=1}^{N}(x_n-\bar x_k)^2$.



\ifCLASSOPTIONcaptionsoff
  \newpage
\fi


\begin{IEEEbiography}[{\includegraphics[width=1in,height=1.25in,clip,keepaspectratio]{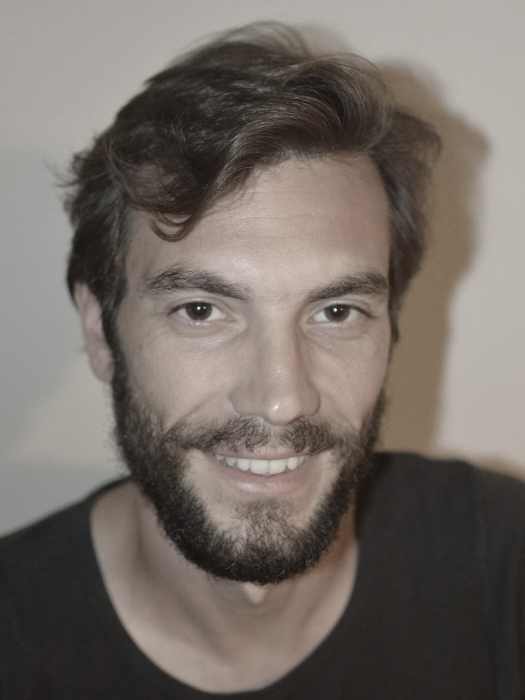}}]{Konstantinos Makantasis}
received his computer engineering diploma from the Technical university of Crete (TUC, Greece) and his Master degree from the same school (DPEM, TUC). His diploma thesis entitled “Human face detection and tracking using AIBO robots”, while his master thesis entitled “Persons’ fall detection through visual cues”. In 2016 Dr. Makantasis received his PhD from the same school working on detection and semantic analysis of object and events through visual cues. He is mostly involved and interested in computer vision, both for visual spectrum (RGB) and hyperspectral data, and in machine learning / pattern recognition and probabilistic programming. He has more than 20 publications in international journals and conferences on computer vision, signal and image processing and machine learning. He has been involved for more than 7 years as a researcher in numerous European and national competing research programs (Interreg, FP7, Marie Curie actions) towards the design, development and validation of state-of-the-art methodologies and cutting-edge technologies in data analytics and computer vision.
\end{IEEEbiography}

\begin{IEEEbiography}[{\includegraphics[width=1in,height=1.25in,clip,keepaspectratio]{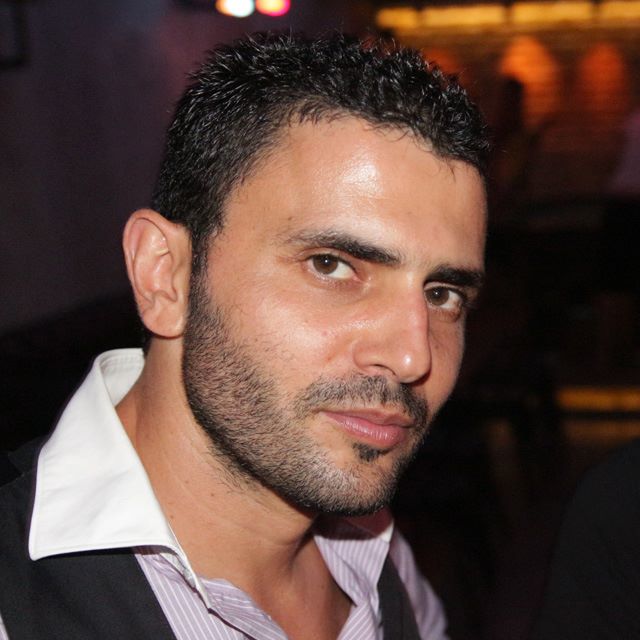}}]{Antonios Nikitakis}
is an engineer and a psychologist. He took his Engineer Diploma in Electrical and Computing Engineering from the Democritus University of Thrace, with specialization in Hardware Computer and Cryptography. He continued his studies in the Technical University of Crete where he received his Master Degree in Electronic and Computer Engineering in which he specialized in Computer Architecture and Hardware Design. In the same institution he fulfilled his Ph.D in Electronic and Computer Engineering with area of specialization the SoC Design in Computer Vision Applications. A part of his research which presented in his Thesis titled:  High Performance Low Power Embedded Vision Systems, rewarded in ESTIMedia 2012 with the Best paper award. At the same time, he graduated with a Bachelor Degree in Psychology from the Psychology Department of the University of Crete. He has years of experience as a Hardware Engineer working in the research and the industry. He has also collaborated with plenty of companies and university to carry out European Programs. Moreover, he has collaborated with experimental psychology labs and he combines his psychology knowledge with his expertise in computer vision and machine learning.
\end{IEEEbiography}

\begin{IEEEbiography}[{\includegraphics[width=1in,height=1.25in,clip,keepaspectratio]{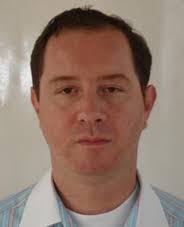}}]
{Anastasios D. Doulamis}
received the Diploma degree in Electrical and Computer Engineering from the National Technical University of Athens (NTUA) in 1995 with the highest honor. In 2001, he has received the PhD degree in video analysis from the NTUA. Until January 2014, he was an associate professor at the Technical University of Crete and now is a faculty member of NTUA. Prof. A. Doulamis has received several awards in his studies, including the Best Greek Student Engineer, Best Graduate Thesis Award, National Scholarship Foundation prize, etc.  He has also served as program committee in several major conferences of IEEE and ACM. He is author of more than 200 papers receiving more than 3000 citations. 
\end{IEEEbiography}

\begin{IEEEbiography}[{\includegraphics[width=1in,height=1.25in,clip,keepaspectratio]{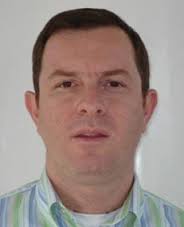}}]
{Nikolaos D. Doulamis}
received a Diploma and PhD in Electrical and Computer Engineering from the National Technical University of Athens (NTUA) both with the highest honor.  He is now Assistant Professor at the NTUA. He has received many awards (e.g., Best Greek Engineer Student, Graduate Thesis Award, NTUA‘s Best Young Medal and best paper awards in IEEE conferences). He has served as Organizer and/or program committee member of major IEEE conferences. He is author of more than 55 (180) journals (conference) papers in the field of signal processing, computational intelligence. He has more than 3000 citations and being involved in European projects. 
\end{IEEEbiography}

\begin{IEEEbiography}[{\includegraphics[width=1in,height=1.25in,clip,keepaspectratio]{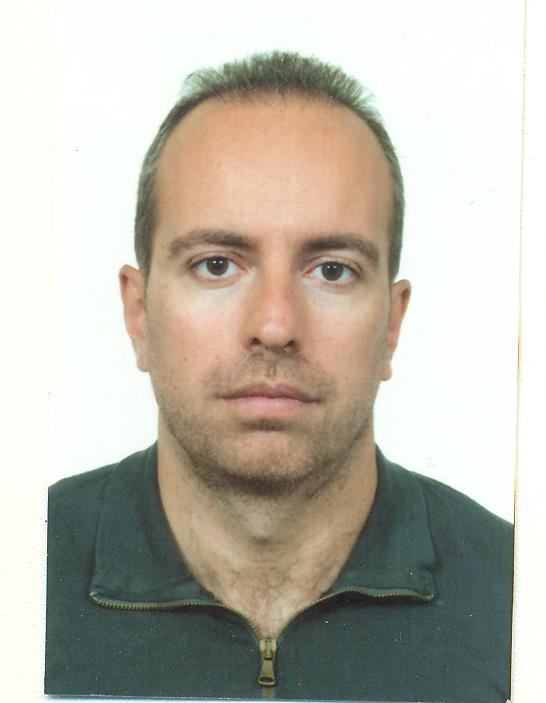}}]
{Ioannis Papaefstathiou}
is a professor in the School of Electronic \& Computer Engineering at the Technical University of Crete. He research interests include the architecture and design of novel computer systems, focusing on devices with highly constrained resources. Papaefstathiou has a PhD in computer science from the University of Cambridge and an M.Sc. from Harvard University.  
\end{IEEEbiography}





\end{document}